\documentclass[10pt,twocolumn,letterpaper]{article}

\usepackage{cvpr}              
\definecolor{cvprblue}{rgb}{0.21,0.49,0.74}
\usepackage[pagebackref,breaklinks,colorlinks,allcolors=cvprblue]{hyperref}

\usepackage[table]{xcolor}
\usepackage{multirow}

\title{Towards Geometry-Aware and Motion-Guided Video Human Mesh Recovery}

\author{
Hongjun Chen,
Huan Zheng,
Wencheng Han,
Jianbing Shen$^{\dagger}$\\
SKL-IOTSC, CIS, University of Macau \\
{\tt\small hj.chen96@outlook.com}, {\tt\small \{huanzheng1998, wencheng256\}@gmail.com}, {\tt\small jianbingshen@um.edu.mo}
}

\begin{document}
\maketitle

\begin{abstract}
Existing video-based 3D Human Mesh Recovery (HMR) methods often produce physically implausible results, stemming from their reliance on flawed intermediate 3D pose anchors and their inability to effectively model complex spatiotemporal dynamics. 
To overcome these deep-rooted architectural problems, we introduce \textbf{HMRMamba}, a new paradigm for HMR that pioneers the use of Structured State Space Models (SSMs) for their efficiency and long-range modeling prowess. 
Our framework is distinguished by two core contributions. 
First, the \textbf{Geometry-Aware Lifting Module}, featuring a novel dual-scan Mamba architecture, creates a robust foundation for reconstruction. It directly grounds the 2D-to-3D pose lifting process with geometric cues from image features, producing a highly reliable 3D pose sequence that serves as a stable anchor. 
Second, the \textbf{Motion-guided Reconstruction Network} leverages this anchor to explicitly process kinematic patterns over time. By injecting this crucial temporal awareness, it significantly enhances the final mesh's coherence and robustness, particularly under occlusion and motion blur. 
Comprehensive evaluations on 3DPW, MPI-INF-3DHP, and Human3.6M benchmarks confirm that HMRMamba sets a new state-of-the-art, outperforming existing methods in both reconstruction accuracy and temporal consistency while offering superior computational efficiency. 
\end{abstract}

\begin{figure}[t]
	\centering
	\includegraphics[width=0.49\textwidth]{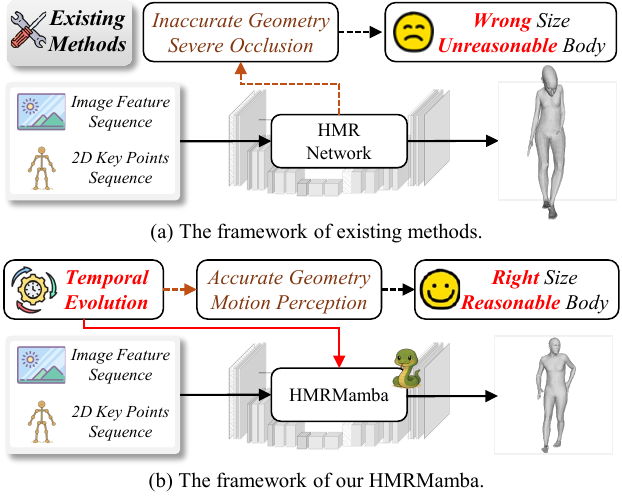} 
	\vspace{-7mm}
	\caption{\textbf{Conceptual comparison of HMR pipelines.} (a) Existing methods often suffer from inaccurate intermediate geometry and poor occlusion handling, leading to flawed reconstructions with wrong body sizes and unnatural poses. (b) Our HMRMamba leverages temporal evolution for accurate geometry perception and motion modeling, resulting in robust and reasonable mesh recovery.}
	\vspace{-5mm}
	\label{fig:intro}  
\end{figure}


\section{Introduction}
\label{sec:intro}

3D Human Mesh Recovery (HMR) from a monocular view is a fundamental task in computer vision, aiming to recover detailed 3D human mesh coordinates from images or videos. This task is both critical and highly challenging for a wide range of applications, including human-computer interaction \cite{sinha2010human}, virtual reality \cite{wohlgenannt2020virtual}, the metaverse \cite{wang2022survey}, and robotics \cite{garcia2007evolution}.

In recent years, significant research has focused on recovering a 3D human mesh from a single image \cite{pose2mesh, cliff, gtrs, simhmr,scorehmr}. These methods typically learn an end-to-end mapping from an input RGB image to the parameters of a statistical body model like SMPL \cite{smpl}. By predicting its pose and shape parameters, a complete 3D human mesh is generated. The SMPL model, specifically, is a parametric representation of the human body defined by 72 pose parameters and 10 shape parameters, which collectively articulate a mesh of 6,890 vertices.

HMR is confronted with inherent challenges, including 2D-to-3D ambiguities, background clutter, and severe occlusions. Since single-image methods often fail to adequately address these problems, leading to unsatisfactory recovery, recent research has increasingly pivoted to video-based HMR. The central premise is to exploit temporally coherent motion in videos to improve mesh reconstruction fidelity. Early video-based approaches \cite{tian2023recovering} extend single-image architectures by extracting per-frame static features with a pre-trained backbone, from which SMPL parameters are directly predicted. However, due to the deficient spatial information and noise inherent in image features, these methods suffer from a difficult trade-off between temporal smoothness and single-frame accuracy, resulting in poor shape fitting and imprecise pose estimates.

Recent advances in video-based 2D Human Pose Estimation (HPE) have achieved high pose accuracy and motion smoothness~\cite{cai2019exploiting,posegcn2,posemamba}. Consequently, current video HMR methods increasingly leverage these skeletal sequences to capture rich spatiotemporal information, typically by integrating 2D-to-3D pose lifting into the HMR pipeline. However, a fundamental representation gap impedes this integration: HPE represents pose with joint coordinates, whereas HMR estimates joint rotations for the SMPL parameters. Bridging this gap remains a significant challenge. PMCE~\cite{pmce} pioneered the introduction of HPE into video HMR, treating joints as auxiliary features fused with image features. While this led to substantial gains, its superficial fusion scheme fails to exploit the distinct advantages of skeletal and image features. To better capitalize on skeletal data, ARTS~\cite{arts} introduces a semi-analytical SMPL regressor. By incorporating human priors and employing disentangled analysis modules, it achieves superior accuracy through a more extensive analytical model.

However, we contend that these approaches suffer from deep-rooted architectural limitations that lead to suboptimal results. As illustrated in Figure~\ref{fig:intro}(a), existing methods often produce physically implausible meshes with incorrect body sizes and unnatural poses, especially under occlusion or complex motion. These failures stem from two fundamental problems.
First, they are built upon a fragile foundation: an intermediate 3D pose anchor. The underlying 2D-to-3D lifting modules, often adapted from HPE techniques, frequently lack sufficient geometric grounding from the image itself. This results in inaccurate anchors that compromise the entire reconstruction process.
Second, they fail to adequately model the intricate kinematic relationships and temporal dynamics of human motion. This deficiency hinders their ability to ensure physical plausibility and temporal coherence, leading to implausible estimations for occluded body parts.

To overcome these fundamental problems, we introduce \textbf{HMRMamba}, a novel framework that, for the first time, pioneers the use of Structured State Space Models (SSMs) for the HMR task, leveraging their efficiency and long-range modeling capabilities. As shown in Figure~\ref{fig:intro}(b), our approach features two synergistic innovations designed to solve the aforementioned issues.
First, to tackle the problem of unreliable anchors, we propose a \textbf{Geometry-Aware Lifting Module}. Powered by our unique \textbf{Spatial Temporal Alignment Mamba (STA-Mamba)} architecture, it directly infuses geometric cues from image features into the lifting process. By leveraging temporal evolution and a dual-scan mechanism, it generates a geometrically-grounded and robust 3D pose sequence, providing a stable anchor for mesh recovery.
Second, to master complex motion dynamics, we introduce the \textbf{Motion-guided Reconstruction Network}. Instead of relying solely on a static pose anchor, this network explicitly processes the temporal dynamics of the 3D joint sequence. This injects crucial kinematic awareness into the final reconstruction, significantly enhancing temporal coherence and robustness against challenging scenarios.

Our contributions are summarized as follows:
\begin{itemize}
    \item \textbf{A New Paradigm for HMR.} We introduce \textbf{HMRMamba}, which pioneers the use of Mamba for HMR. It leverages temporal evolution to address the long-standing issues of monocular ambiguity and occlusion, setting a new direction for the field.

    \item \textbf{Geometry-Aware Lifting Module.} We propose a novel Geometry-Aware Lifting Module with a unique dual-scan Mamba architecture. It produces geometrically-grounded 3D pose anchors, solving the critical problem of unreliable 3D poses that plagues prior work.

    \item \textbf{Motion-guided Reconstruction Network.} We introduce a novel Motion-guided Reconstruction Network that injects kinematic awareness into the mesh recovery process. This ensures temporally coherent and physically plausible reconstructions.

    \item \textbf{State-of-the-Art Performance.} Extensive experiments demonstrate that HMRMamba establishes a new state-of-the-art across major benchmarks (3DPW, MPI-INF-3DHP, and Human3.6M) with computational efficiency.
\end{itemize}

\section{Related Work}

\subsection{3D Human Mesh Recovery}
Due to the wide accessibility of images, most human mesh recovery methods use a single image as input. These approaches generally fall into two categories: parametric and non-parametric. Parametric methods estimate body model parameters based on human models like SMPL~\cite{smpl}, often incorporating priors such as part segmentation~\cite{pare}, bounding boxes~\cite{cliff}, or kinematic constraints~\cite{hkmr,hybrik,gator}, sometimes leveraging 3D pose information for guidance. Non-parametric methods directly predict mesh vertices from images or 2D poses, using networks such as multi-stage upsampling~\cite{pose2mesh} or graph transformers~\cite{gtrs} to reconstruct the mesh.
Although image-based methods achieve high accuracy, they often generate unsmooth motions when applied to videos. In contrast, video-based HMR methods aim for both accuracy and temporal consistency by designing temporal modeling networks. Early works utilize GRU-based models~\cite{vibe,meva,tcmr} to smooth motion, but struggle with long-term dependencies. To address this, most recent approaches adopt Transformer-based networks~\cite{mead,glot,bicf,unspat} to better capture temporal correlations and spatiotemporal structure. However, their performance remains limited by insufficient spatial information and noisy image features. Moreover, while skeleton priors are commonly used in image-based HMR, they are often overlooked in video-based HMR. Some recent methods, such as Sun~\textit{et al.}\cite{r4-1} and PMCE\cite{pmce}, attempt to leverage skeleton information, but either do not fully utilize structural relationships or ignore skeleton structure during feature fusion. ARTS~\cite{arts} further disentangles the joints and initializes SMPL with priors to obtain more diverse SMPL initialization parameters, thereby further improving mesh accuracy.

\subsection{Mamba Models for Vision Tasks}
Recent years have seen renewed interest in State Space Models (SSMs) for sequence modeling, due to their ability to efficiently capture long-range dependencies with lower computational costs~\cite{gu2023mamba}. Building on this foundation, Mamba and Mamba2 introduce hardware-friendly parallel algorithms for linear-time inference and improved parameter efficiency, outperforming Transformer-based architectures in both speed and scalability~\cite{gu2023mamba, mamba2}.
Recently, Mamba has been adopted for constructing vision foundation models~\cite{zhu2024vision, liu2024vmamba, guo2025mambair, mobilemamba} and introduces various scanning mechanisms tailored to the characteristics of 2D images or 3D videos, including bi-directional scan~\cite{zhu2024vision}, cross-scan~\cite{liu2024vmamba}, local scan~\cite{huang2024localmamba, posemamba, rawmamba}.
It is noted that Mamba offer strong potential for human-centric tasks thanks to their efficient global modeling. Recent works have advanced the field by enhancing temporal and spatial modeling (Motion Mamba~\cite{motionmamba}), incorporating graph learning for structured joint relations (Hamba~\cite{hamba}), and improving spatiotemporal joint dependency modeling for 2D-to-3D pose lifting (PoseMamba~\cite{posemamba}; Posemagic~\cite{posemagic}), where hybrid approaches adaptively combine Mamba with GCNs. 
To the best of our knowledge, the Mamba architecture has not been applied to the HMR domain, where the characteristics of the mesh domain differ significantly from those of traditional computer vision tasks. Therefore, this paper proposes a novel Mamba structure specifically designed with a focus on mesh feature.

\begin{figure*}[t]
    \centering
    \includegraphics[width=0.95\linewidth]{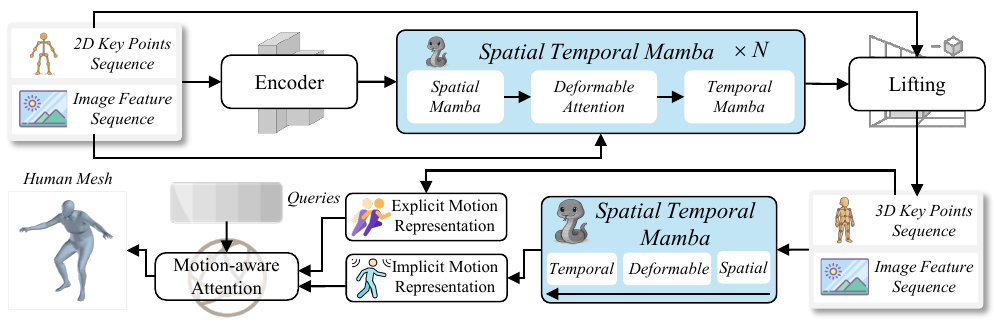}
    \vspace{-2mm}
    \caption{
    \textbf{The two-stage architecture of HMRMamba. }
    \textbf{(1) Geometry-Aware Lifting Module (top row):} To address the problem of unreliable anchors, our novel \textbf{STA-Mamba} infuses geometric cues from image features into the lifting process. This produces a robust, geometrically-grounded 3D pose sequence that serves as a stable anchor.
    \textbf{(2) Motion-guided Reconstruction Network (bottom row):} Leveraging this stable anchor, a motion-aware attention mechanism models kinematic dynamics. This injects temporal awareness into the final regression to ensure a coherent and physically plausible mesh.
    }
    \vspace{-6mm}
    \label{fig:arch}
\end{figure*}

\section{Method}

\subsection{Preliminaries}
SSMs are inspired by continuous linear time-invariant systems and are designed to efficiently map one-dimensional sequences $x(t) \in \mathbb{R} \mapsto y(t) \in \mathbb{R}$ via a hidden state $h(t) \in \mathbb{R}^\mathtt{N}$, where $\mathtt{N}$ is the size of hidden state. SSMs model the input data by employing the following Ordinary Differential Equation (ODE):
\begin{equation}
    \begin{aligned}
        \label{eq:lti}
        h'(t) &= \mathbf{A}h(t) + \mathbf{B}x(t), \\
        y(t) &= \mathbf{C}h(t),
    \end{aligned}
\end{equation}
where $\mathbf{A} \in \mathbb{R}^{\mathtt{N} \times \mathtt{N}}$ denotes the evolution parameter and $\mathbf{B} \in \mathbb{R}^{\mathtt{N} \times 1}$, $\mathbf{C} \in \mathbb{R}^{1 \times \mathtt{N}}$ denotes the projection parameters.

Mamba~\cite{gu2023mamba} provides a discrete approximation to the underlying continuous-time system by employing the Zero-Order Hold (ZOH) approach, which introduces a timescale parameter $\mathbf{\Delta}$. This parameter is used to map the continuous matrices $\mathbf{A}$ and $\mathbf{B}$ to their discrete counterparts, $\mathbf{\overline{A}}$ and $\mathbf{\overline{B}}$, as follows:
\begin{equation}
    \begin{aligned}
        \label{eq:zoh}
        \mathbf{\overline{A}} &= \exp(\mathbf{\Delta} \mathbf{A}), \\
        \mathbf{\overline{B}} &= (\mathbf{\Delta}\mathbf{A})^{-1}\big(\exp(\mathbf{\Delta} \mathbf{A}) - \mathbf{I}\big)\mathbf{\Delta}\mathbf{B}.
    \end{aligned}
\end{equation}

With these discrete parameters, the original continuous linear time-invariant (LTI) system can be reformulated at each time step using a step size of $\mathbf{\Delta}$:
\begin{equation}
    \begin{aligned}
        \label{eq:discrete_lti}
        h_t &= \mathbf{\overline{A}}h_{t-1} + \mathbf{\overline{B}}x_t,\quad y_t = \mathbf{C}h_t.
    \end{aligned}
\end{equation}

Moreover, this discrete recurrence admits an efficient, parallelizable implementation via a global convolution across the input sequence:
\begin{equation}
    \begin{aligned}
        \label{eq:conv}
        \mathbf{y} = \mathbf{x} \circledast \mathbf{\overline{K}},
    \end{aligned}
\end{equation}
where the convolution kernel is defined as
\begin{equation*}
    \mathbf{\overline{K}} = \left(\mathbf{C}\mathbf{\overline{B}},\  \mathbf{C}\mathbf{\overline{A}}\mathbf{\overline{B}},\  \ldots,\  \mathbf{C}\mathbf{\overline{A}}^{\mathtt{L}-1}\mathbf{\overline{B}}\right),
\end{equation*}
with $\mathtt{L}$ denoting the sequence length, $\circledast$ the convolution operator, and $\mathbf{\overline{K}}\in\mathbb{R}^{\mathtt{L}}$ the resulting kernel. This convolutional formulation enables Mamba to efficiently model long-range dependencies over sequences.

\subsection{Human Mesh Recovery Pipeline}
The overall architecture of our proposed HMRMamba, illustrated in Figure~\ref{fig:arch}, is a two-stage framework meticulously designed to transform a monocular video sequence into a temporally coherent and accurate 3D human mesh. The pipeline systematically addresses the core challenges of existing methods by first establishing a robust geometric foundation and then refining the mesh with motion-aware visual context.

Given an input video sequence $\mathcal{V} = \{I_t\}_{t=1}^T$ consisting of $T$ frames, we begin by extracting essential low-level features. A pre-trained ResNet-50 serves as the image backbone to extract a sequence of visual features $F_{\text{img}} \in \mathbb{R}^{T \times D}$, where $D=2048$. Concurrently, a 2D pose detector is employed to estimate the 2D joint locations for each frame, yielding a 2D pose sequence $P_{\mathrm{2D}} \in \mathbb{R}^{T \times J \times 2}$, where $J$ is the number of body joints.

The first stage of our pipeline is the \textit{Geometry-Aware Lifting Module}, whose primary function is to overcome the prevalent issue of unreliable intermediate 3D pose anchors. This module, denoted as $\Phi_{\text{lift}}$, takes the 2D pose sequence $P_{\mathrm{2D}}$ and image features $F_{\text{img}}$ as input, and generates a geometrically-grounded and temporally stable 3D pose sequence $P_{\mathrm{3D}}$. This process can be formulated as:
\begin{equation}
    P_{\mathrm{3D}} = \Phi_{\text{lift}}(P_{\mathrm{2D}}, F_{\text{img}}),
    \label{eq:lifting}
\end{equation}
where $P_{\mathrm{3D}} \in \mathbb{R}^{T \times J \times 3}$. As will be detailed in Section~\ref{sec:lifting_module}, the core of this module is our novel STA-Mamba architecture, which explicitly models both temporal dynamics and kinematic constraints to ensure the anatomical plausibility of the generated 3D skeleton.

The second stage is the \textit{Motion-guided Reconstruction Network}. Instead of relying on a static, single-frame anchor, this stage leverages the entire 3D pose sequence $P_{\mathrm{3D}}$ to guide the final mesh regression. This network, denoted as $\Psi_{\text{recon}}$, utilizes the rich kinematic information embedded in $P_{\mathrm{3D}}$ to interpret visual cues from $F_{\text{img}}$ more effectively, particularly in challenging scenarios like occlusion and motion blur. The output is the final 3D mesh vertex sequence $V_{\text{mesh}}$, regressed as:
\begin{equation}
    V_{\mathrm{mesh}} = \Psi_{\text{recon}}(P_{\mathrm{3D}}, F_{\text{img}}),
    \label{eq:reconstruction}
\end{equation}
where $V_{\text{mesh}} \in \mathbb{R}^{T \times N \times 3}$, and $N$ represents the 6,890 vertices of the SMPL model. This motion-guided method ensures that the final mesh is not only accurate in shape but also exhibits physically plausible and coherent motion.

\begin{figure*}[t]
    \centering
    \includegraphics[width=0.8\linewidth]{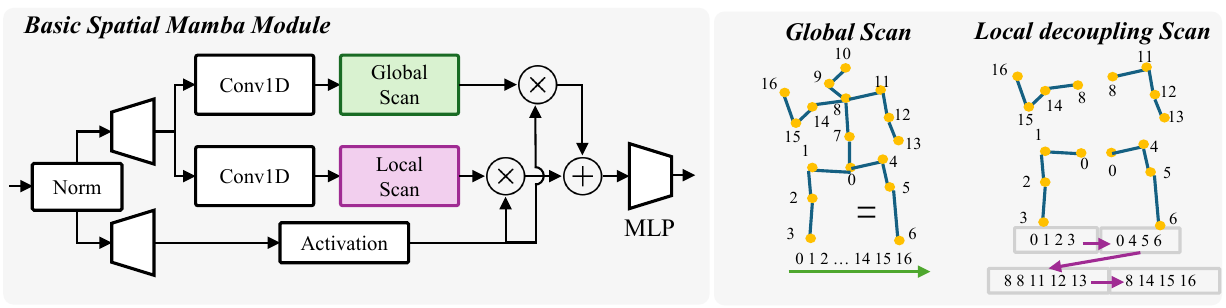}
    \vspace{-2mm}
    \caption{\textbf{Detailed architecture of our Dual-Scan Mamba Block.} 
    The \textbf{Global Scan} branch processes the sequence linearly to capture holistic, long-range dependencies. In contrast, the \textbf{Local Scan} branch employs a non-sequential scanning order that follows the human kinematic tree, as illustrated on the right. This novel local scan explicitly models anatomical constraints by traversing joints along natural limbs. The outputs of these two complementary branches are fused via element-wise multiplication and addition, producing a global context and local structural plausibility representation. 
    }
    \vspace{-6mm}
    \label{fig:mamba}
\end{figure*}

\subsection{Geometry-Aware Lifting Module}
\label{sec:lifting_module}

Existing video-HMR methods often rely on an intermediate 3D pose anchor generated by lifting a 2D pose sequence. However, we contend that this 2D-to-3D lifting process is often geometrically agnostic, prioritizing overall pose plausibility at the cost of local kinematic details like bone lengths or joint articulation. This results in unreliable 3D anchors that compromise the final mesh reconstruction. To address this, we introduce the \textbf{Geometry-Aware Lifting Module}, whose architecture is depicted in the top row of Figure~\ref{fig:arch}. This module is designed to produce a robust and anatomically plausible 3D pose sequence by explicitly modeling spatial structure and temporal dynamics.

The module operates in a sequential pipeline. First, an \textbf{Encoder} fuses the input 2D pose sequence $P_{\mathrm{2D}}$ and image features $F_{\text{img}}$ into an initial joint feature representation. This representation is then processed by our novel \textbf{STA-Mamba} (Spatial-Temporal Alignment Mamba) architecture, which consists of two main stages: spatial modeling and temporal modeling.

\paragraph{Spatial Mamba for Intra-frame Structure.}
The \textbf{Spatial Mamba} block operates on a per-frame basis to enforce intra-frame anatomical correctness. It refines the spatial relationships between joints within each frame, ensuring the pose is structurally sound. This stage produces a spatially-refined feature sequence, $F_{\text{spatial}}$, formulated as:
\begin{equation}
    F_{\text{spatial}} = \text{SpatialMamba}(\text{Encoder}(P_{\mathrm{2D}}, F_{\text{img}})).
    \label{eq:spatial_mamba}
\end{equation}

\paragraph{Temporal Mamba for Inter-frame Dynamics.}
Subsequently, a \textbf{Deformable Attention} mechanism is employed. It uses the image feature query at each location $i$ to predict sampling offsets $\Delta \mathbf{p}$ and attention weights $A$. These are then used to sample and aggregate features 2D key points or 3D key points $\mathbf{v}$. The spatial feature $F'_{\text{spatial}}$ at location $i$ is formulated as:
\begin{equation}
\label{eq:deformable_attn}
F^{\prime}_{\text{spatial}}[i] = \sum_{m=1}^{M} \mathbf{W}_m \left[ \sum_{k=1}^{K} A_{mik} \cdot \mathbf{W}'_m \mathbf{v}\left( p_i + \Delta \mathbf{p}_{mik} \right) \right]
\end{equation}
where $M$ is the number of attention heads, $K$ is the number of sampled points per head, and $i$ is the query location. The offsets $\Delta \mathbf{p}_{mik}$ and attention weights $A_{mik}$ are predicted from the query feature $\mathbf{q}_i$. The term $\mathbf{v}(p_i + \Delta \mathbf{p}_{mik})$ represents the value feature sampled from $F_{\text{img}}$ at the refined 3D location $p_i + \Delta \mathbf{p}_{mik}$. $\mathbf{W}_m$ and $\mathbf{W}'_m$ are learnable weight matrices to dynamically align rich visual cues from the original image features $F_{\text{img}}$ with the now geometrically-refined pose representation $F_{\text{spatial}}$. This provides crucial appearance context for handling ambiguities like self-occlusion. The enriched features are then passed to the \textbf{Temporal Mamba} block, which processes the sequence across the time dimension. It models the inter-frame evolution of each joint's representation, ensuring the resulting motion is both smooth and coherent. This step yields the final, spatio-temporally aware feature sequence, 
\begin{equation}
    F_{\text{temporal}} = \text{TemporalMamba}( F^{\prime}_{\text{spatial}} ).
    \label{eq:temporal_mamba}
\end{equation}

\paragraph{The Dual-Scan Mamba Block.}
The key innovation powering both our Spatial and Temporal Mamba blocks is the novel \textbf{Dual-Scan Mamba Block}, detailed in Figure~\ref{fig:mamba}. It enhances the standard Mamba with a parallel, dual-scan mechanism to capture both holistic and structured relationships. Given an input feature set, it performs two scans concurrently:
\begin{itemize}
    \item \textbf{Global Scan:} This scan processes the input elements in a linear, sequential order (e.g., by joint index or time step). It serves to capture the overall, long-range dependencies within the sequence.
    \item \textbf{Local Scan (Kinematic Scan):} This is our core contribution for injecting geometric priors. This scan abandons the linear order and instead traverses the elements along the human kinematic tree (e.g., torso-shoulder-elbow-wrist), explicitly encoding anatomical constraints into the learned representation.
\end{itemize}
The outputs of these two complementary scans, denoted as $O_{\text{global}}$ and $O_{\text{local}}$, are fused to produce a powerful representation $O_{\text{fused}}$ that is aware of both global context and local structure. This fusion process can be described as:
\begin{equation}
    O_{\text{fused}} = \text{Activation}(\text{Conv1D}(O_{\text{global}})) \odot O_{\text{local}},
    \label{eq:dual_scan_fusion}
\end{equation}
where $\odot$ denotes element-wise multiplication and the activation function is typically Sigmoid-weighted Linear Unit (SiLU). This dual-scan mechanism is adeptly specialized for each stage: in the Spatial Mamba, the Kinematic Scan traverses the skeletal tree of each pose, while in the Temporal Mamba, it processes temporal dynamics along kinematic chains.

\paragraph{Lifting Head.}
Finally, a simple multi-layer perceptron (MLP) acts as the \textbf{Lifting Head}. It takes the comprehensive feature representation $F_{\text{temporal}}$ and regresses it into the final 3D joint coordinates, producing a highly reliable 3D pose sequence $P_{\mathrm{3D}}$ that serves as a stable anchor for the subsequent mesh reconstruction. This process is defined as:
\begin{equation}
    P_{\mathrm{3D}} = \text{Lifting}(F_{\text{temporal}}).
    \label{eq:lifting_head}
\end{equation}
This robustly grounded 3D pose ensures both geometric and temporal reality for the next stage.

\subsection{Motion-guided Reconstruction Network}
\label{sec:reconstruction_network}

Previous methods often utilize only a single-frame 3D pose to guide mesh recovery. However, relying solely on frame-wise predictions introduces considerable uncertainty, especially in challenging scenarios like occlusion or rapid motion, where the 3D pose anchor can be unreliable and degrade mesh accuracy. To overcome this limitation, our framework leverages the full temporal context available from the preceding stage.

The \textbf{Motion-guided Reconstruction Network}, illustrated in the bottom row of Figure~\ref{fig:arch}, is designed to capitalize on the rich kinematic information embedded in the entire 3D pose sequence, $P_{\mathrm{3D}} \in \mathbb{R}^{T \times J \times 3}$, generated by our Geometry-Aware Lifting Module. This sequence serves as a dynamic, motion-aware anchor, providing robust temporal guidance for the final mesh regression.
The core of this network is a \textbf{Motion-aware Attention} mechanism. This module explicitly models the relationship between the motion of the 3D skeleton and the per-frame visual features. It formulates queries from the image features $F_{\text{img}}$ and uses the 3D pose sequence $P_{\mathrm{3D}}$ to construct both keys and values. This allows the network to selectively focus on visual evidence that is consistent with the established 3D motion, effectively filtering out noise and resolving ambiguities.

Specifically, we derive two distinct motion representations from the pose sequence $P_{\mathrm{3D}}$:
\begin{itemize}
    \item \textbf{Explicit Motion Representation ($M_{\text{exp}}$):} This is computed as the frame-to-frame displacement of the 3D joints, representing the velocity. It is formulated as $M_{\text{exp}}^{(t)} = P_{\mathrm{3D}}^{(t)} - P_{\mathrm{3D}}^{(t-1)}$ for frame $t$. This provides explicit kinematic cues.
    \item \textbf{Implicit Motion Representation ($M_{\text{imp}}$):} This is achieved by using motion information from $P_{\mathrm{3D}}$ to correct the image sequence features, which implicitly encodes posture over time.
\end{itemize}
These motion representations are then projected and used to generate attention keys ($K$) and values ($V$), while the image features $F_{\text{img}}$ are used to generate queries ($Q$). The motion-aware attention output, $F_{\text{motion-aware}}$, is computed as:
\begin{equation}
    F_{\text{motion-aware}} = \text{Attention}(Q, K, V) = \text{softmax}\left(\frac{QK^T}{\sqrt{d_k}}\right)V,
    \label{eq:motion_attention}
\end{equation}
where $Q = W_Q F_{\text{img}}$, and $K, V$ are derived from linear projections of both $M_{\text{exp}}$ and $M_{\text{imp}}$.

The resulting motion-aware features $F_{\text{motion-aware}}$ are then processed by a final regression head, $\mathcal{R}_{\text{recon}}$, typically an MLP or a simple decoder, to predict the final 3D mesh vertex sequence $V_{\text{mesh}} \in \mathbb{R}^{T \times N \times 3}$, where $N=6890$ is the number of vertices in the SMPL model. The overall reconstruction process is formulated as:
\begin{equation}
    V_{\text{mesh}} = \mathcal{R}_{\text{recon}}(F_{\text{motion-aware}}).
    \label{eq:mesh_reconstruction}
\end{equation}

By explicitly conditioning the reconstruction on the temporal dynamics of the 3D joints, this network ensures that the final mesh is not only accurate in shape for each frame but also exhibits physically plausible and coherent motion over time. This approach significantly enhances robustness against challenges like occlusion, where single-frame visual information might be insufficient.

\subsection{Loss Function}
For 3D pose lifting, we adopt a multi-term loss function, following~\cite{motionbert}, that combines four objectives: (1) the Mean Per Joint Position Error (MPJPE) loss $\mathcal{L}_{3D}$ for measuring 3D joint accuracy, (2) a temporal consistency loss $\mathcal{L}_t$, (3) the Mean Per Joint Velocity Error (MPJVE) loss $\mathcal{L}_m$ for temporal smoothness, and (4) a 2D re-projection loss $\mathcal{L}_{2D}$ that aligns predicted 2D joint projections with image observations~\cite{motionbert}:
\begin{equation}
    \mathcal{L}_{pose} = \mathcal{L}_{3D} + \lambda_t \mathcal{L}_t + \lambda_m \mathcal{L}_m + \lambda_{2D} \mathcal{L}_{2D},
\end{equation}
where $\lambda_t=0.5$, $\lambda_m=20$ and $\lambda_{2D}=0.5$. 


For mesh recovery supervision, we introduce a set of four loss terms: mesh vertex loss $\mathcal{L}_{\text{mesh}}$, 3D joint loss $\mathcal{L}_{\text{joint}}$, surface normal loss $\mathcal{L}_{\text{normal}}$, and surface edge loss $\mathcal{L}_{\text{edge}}$. The overall mesh recovery loss is formulated as a weighted sum:
\begin{equation}
    \mathcal{L}_{\text{mesh\_recovery}} = \lambda_{m} \mathcal{L}_{\text{mesh}} + \lambda_{j} \mathcal{L}_{\text{joint}} + \lambda_{n} \mathcal{L}_{\text{normal}} + \lambda_{e} \mathcal{L}_{\text{edge}},
\end{equation}
where the weights are empirically set to $\lambda_{m}=1$, $\lambda_{j}=1$, $\lambda_{n}=0.1$, and $\lambda_{e}=20$.


\begin{table*}[tbph]
\caption{Evaluation of HMR methods on 3DPW, MPI-INF-3DHP, and Human3.6M datasets.  \textbf{Bold}: best. \underline{Underline}: suboptimal}
\vspace{-5mm}
\begin{center}
\small
\renewcommand\arraystretch{1.25}  
\setlength{\tabcolsep}{0.8mm}{
\begin{tabular}{l|cccc|ccc|ccc}
\toprule
\multirow{2}{*}{Method} & \multicolumn{4}{c|}{3DPW} & \multicolumn{3}{c|}{MPI-INF-3DHP}      & \multicolumn{3}{c}{Human3.6M} \\  \cline{2-11}
      & {\scriptsize MPJPE~$\downarrow$}
      & {\scriptsize PA-MPJPE~$\downarrow$}
      & {\scriptsize MPVPE~$\downarrow$}
      & {\scriptsize Accel~$\downarrow$}
      & {\scriptsize MPJPE~$\downarrow$}
      & {\scriptsize PA-MPJPE~$\downarrow$}
      & {\scriptsize Accel~$\downarrow$}
      & {\scriptsize MPJPE~$\downarrow$}
      & {\scriptsize PA-MPJPE~$\downarrow$}
      & {\scriptsize Accel~$\downarrow$} \\ 
      \hline \hline
{\footnotesize TCMR (CVPR'21)}  \cite{tcmr} & 86.5  & 52.7 & 102.9 & 6.8   & 97.3  & 63.5 & 8.5   & 73.6 & 52.0  & 3.9  \\ 
{\footnotesize MEAD (ICCV'21)} \cite{mead} & 79.1 &  \underline{45.7}   & 92.6  & 17.6  & 83.6 & 56.2  & -     & 56.4 & 38.7  & -    \\
{\footnotesize MPS-Net (CVPR'22)} \cite{mpsnet} & 84.3 & 52.1   & 99.7  & 7.6  & 96.7 & 62.8   & 9.6   & 69.4 & 47.4  & \underline{3.6}  \\
{\footnotesize Zhang (CVPR'23)} \cite{zhang}  & 83.4  & 51.7  & 98.9  & 7.2   & 98.2  & 62.5 & 8.6  & 73.2  & 51.0  & \underline{3.6}  \\
{\footnotesize GLoT (CVPR'23)} \cite{glot}  & 80.7  & 50.6  & 96.3  & 6.6  & 93.9  & 61.5  & 7.9  & 67.0  & 46.3  & \underline{3.6}  \\
{\footnotesize Bi-CF (MM'23)} \cite{bicf} & 73.4  & 51.9  & 89.8  & 8.8  & 95.5  & 62.7  & 7.7  & 63.9 & 46.1   &  \textbf{3.1} \\
{\footnotesize PMCE (ICCV'23)} \cite{pmce} & 69.5  & {46.7}  & 84.8  & \textbf{6.5}  &  79.7  & 54.5 & \textbf{7.1}   &  53.5 &  37.7 & \textbf{3.1} \\
{\footnotesize UNSPAT (WACV'24)} \cite{unspat} & 75.0  & \textbf{45.5}  & 90.2  & 7.1   & 94.4  & 60.4 & 9.2   & 58.3 & 41.3  & 3.8  \\
{\footnotesize ARTS (MM'25)} \cite{arts} & \underline{67.7} & 46.5 & 81.4 & \textbf{6.5} & 71.8  & 53.0 & 7.4   & 51.6 & 36.6  & \textbf{3.1}   
\\ 
\hline \hline
{\footnotesize Ours-S}  & 66.9 & 46.3 & \underline{81.4} & \underline{6.6} & \underline{70.1}  & \underline{51.7} & { \underline{7.2} }    & \underline{51.2} & \underline{36.0}  & {\textbf{3.1} }   \\

{\footnotesize Ours-L}  & \textbf{64.8} & \textbf{45.5} & \textbf{79.8} & \textbf{6.5} & \textbf{68.3}  & \textbf{50.2} &  \textbf{7.1}  & \textbf{49.3} & \textbf{35.7}  & {\textbf{3.1} }   
\\
\bottomrule
\end{tabular}}
\label{table:tab1}
\end{center}
\vspace{-6mm}
\end{table*}

\section{Experiments}

\subsection{Experiments setup}

\noindent
\textbf{Datasets.} We evaluate our method on three widely-used benchmarks: 3DPW, MPI-INF-3DHP, and Human3.6M.

\noindent
\textbf{3DPW}~\cite{3dpw} is a challenging in-the-wild dataset providing 3D mesh annotations. We use its official training and testing splits for experiments on this benchmark.
\textbf{MPI-INF-3DHP}~\cite{mpii3d} is a large-scale 3D pose dataset with diverse scenes. It provides accurate 3D joint annotations from a marker-less motion capture system. We report results on the official test set to evaluate generalization.
\textbf{Human3.6M}~\cite{h36m} is a large-scale dataset from a controlled lab setting. We use the standard protocol, training on subjects S1, S5, S6, S7, and S8, and testing on S9 and S11, with evaluation performed on the ground-truth 3D joint locations.

\noindent
\textbf{Evaluation metrics.}
To evaluate per-frame accuracy, we employ the mean per joint position error (MPJPE), Procrustes-aligned MPJPE (PA-MPJPE), and mean per vertex position error (MPVPE). These metrics measure the difference between the predicted mesh position and ground truth in millimeters ($mm$). To evaluate temporal consistency, we utilize the acceleration error (Accel) proposed in HMMR~\cite{hmmr}. This metric calculates the average difference in acceleration of joints, which is measured in $mm/s^2$.

\noindent
\textbf{Implements details.}
Following prior video-based approaches~\cite{pmce, arts}, we employ ResNet-50 pretrained by SPIN~\cite{spin} as our static image feature extractor. For 2D pose estimation, we use CPN~\cite{cpn} on Human3.6M and ViTPose~\cite{vitpose} on 3DPW and MPI-INF-3DHP. Input sequences are constructed with 16 frames for fair comparison, sampled with a stride of 4. The pose lifting stage is optimized using Adam~\cite{kingma2014adam} with an initial learning rate of $2\times10^{-4}$, a batch size of 64, and trained for 100 epochs. Weight decay is set to 0.01, and the learning rate is multiplicatively reduced by 0.99 each epoch. In the mesh recovery stage, we initialize the model with pre-trained weights from the 3D pose estimation stream and fine-tune the entire network for 20 epochs using a batch size of 32 and a learning rate of $5\times 10^{-5}$. The 3D pose stream is configured with 3 layers and a feature dimension of 256, while the co-evolution block comprises 3 layers with a feature dimension of 64.
All experiments are conducted on an RTX 4090 GPU.

\subsection{Main Results}

We provide a comprehensive quantitative comparison of our proposed method, HMRMamba, with several state-of-the-art approaches on the several benchmarks. Results are summarized in~\cref{table:tab1}, where Ours-S denotes a Mamba block depth of 3 and Ours-L denotes a depth of 5.

\noindent
\textbf{Performance on Human3.6M.} On the widely-used Human3.6M benchmark, our method achieves an MPJPE of 51.2 mm and a PA-MPJPE of 36.0 mm, surpassing previous leading methods like ARTS (51.6 mm) and PMCE (53.5 mm). Crucially, HMRMamba also achieves the best temporal consistency, with an acceleration error (Accel) of 3.1 mm/s², matching the SOTA performance while significantly improving pose accuracy. This highlights our model's ability to produce both accurate and temporally smooth mesh.

\noindent
\textbf{Performance on 3DPW.} For the challenging in-the-wild 3DPW dataset, HMRMamba demonstrates remarkable performance, reducing the MPJPE to 66.9 mm, a significant improvement over the prior best (67.7 mm from ARTS). Furthermore, it achieves the best MPVPE of 81.4 mm (tying with ARTS) and a highly competitive acceleration error of 6.6 mm/s². This underscores our model's robustness to complex outdoor scenes, varied clothing, and occlusions.

\noindent
\textbf{Performance on MPI-INF-3DHP.} To evaluate generalization, we test on MPI-INF-3DHP. Our model achieves an MPJPE of \textbf{70.1 mm} and a PA-MPJPE of 51.7 mm, outperforming all competitors by a substantial margin. This result validates the effectiveness of our approach in generalizing to subjects and environments not seen during training. 

\begin{figure*}[t]
	\centering
	{%
        \begin{minipage}{0.9\textwidth}
            \centering
            \footnotesize
            \includegraphics[width=0.19\textwidth,height=0.12\textwidth]{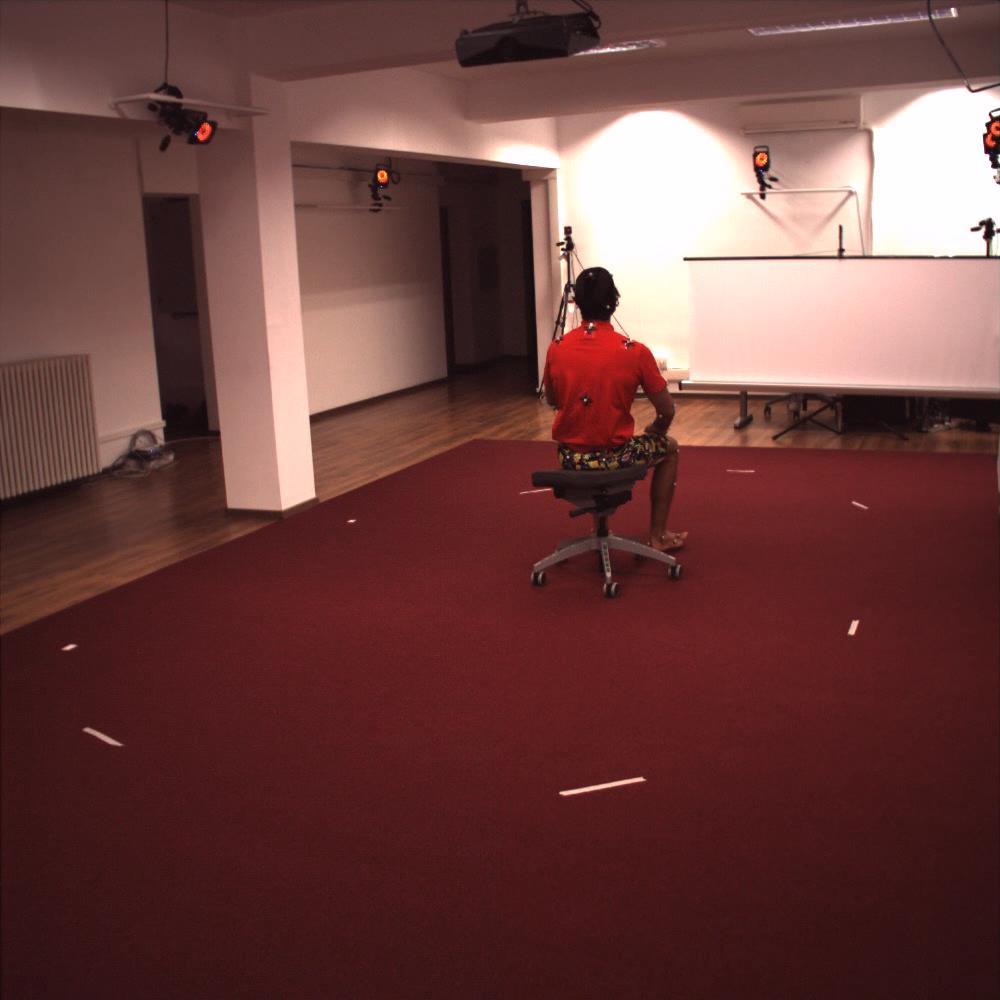}
            \includegraphics[width=0.19\textwidth,height=0.12\textwidth]{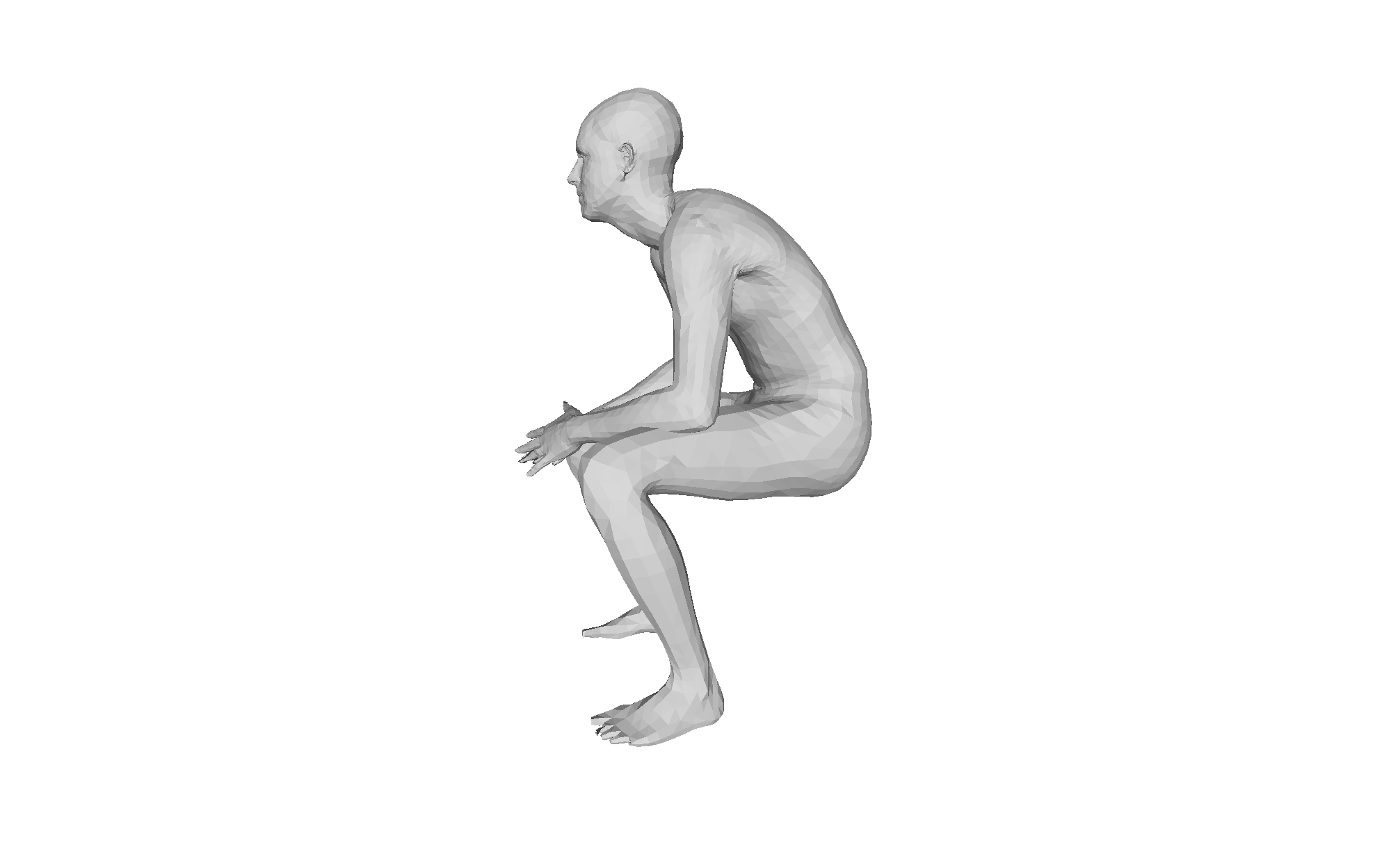}
            \includegraphics[width=0.19\textwidth,height=0.12\textwidth]{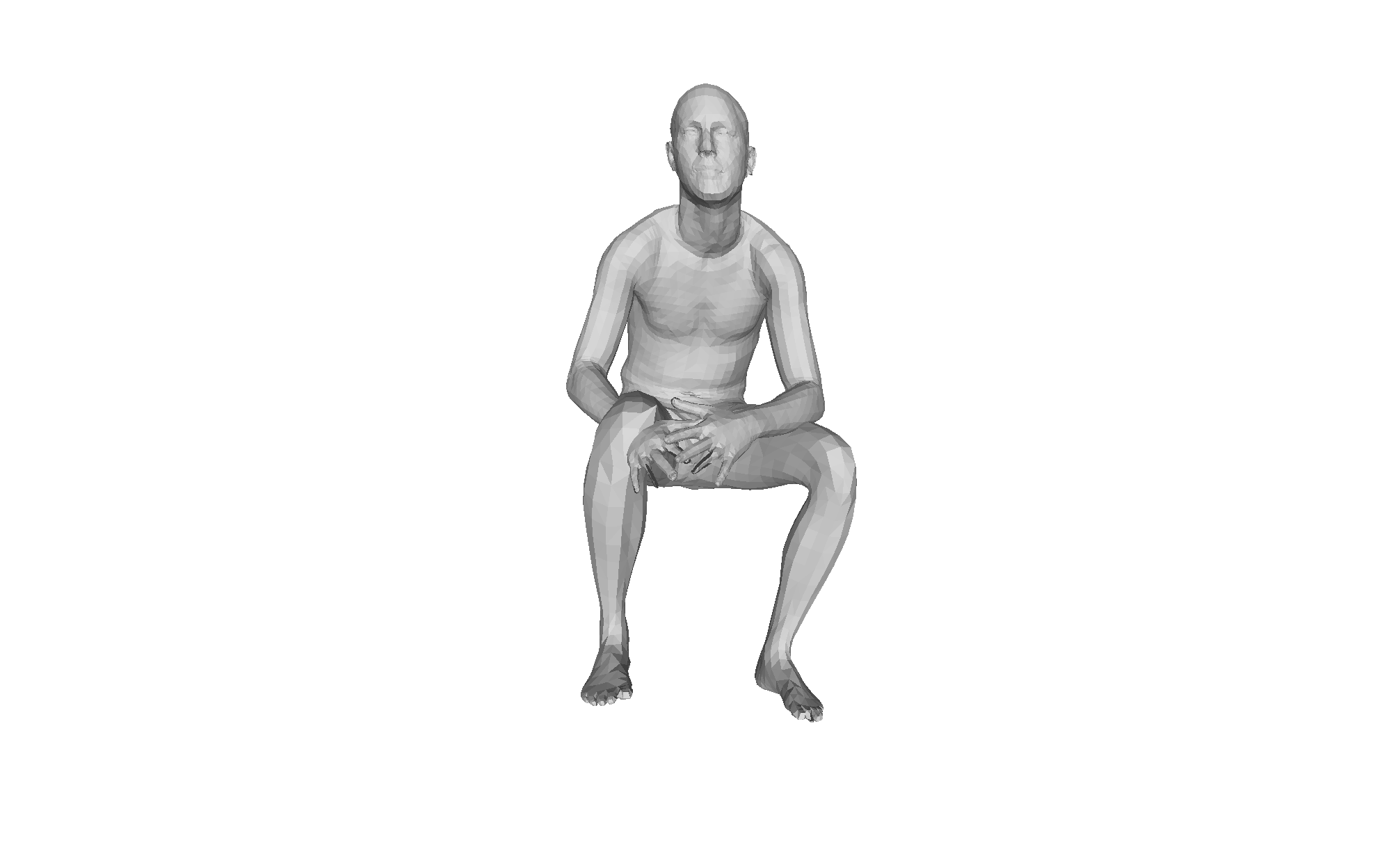}
            \includegraphics[width=0.19\textwidth,height=0.12\textwidth]{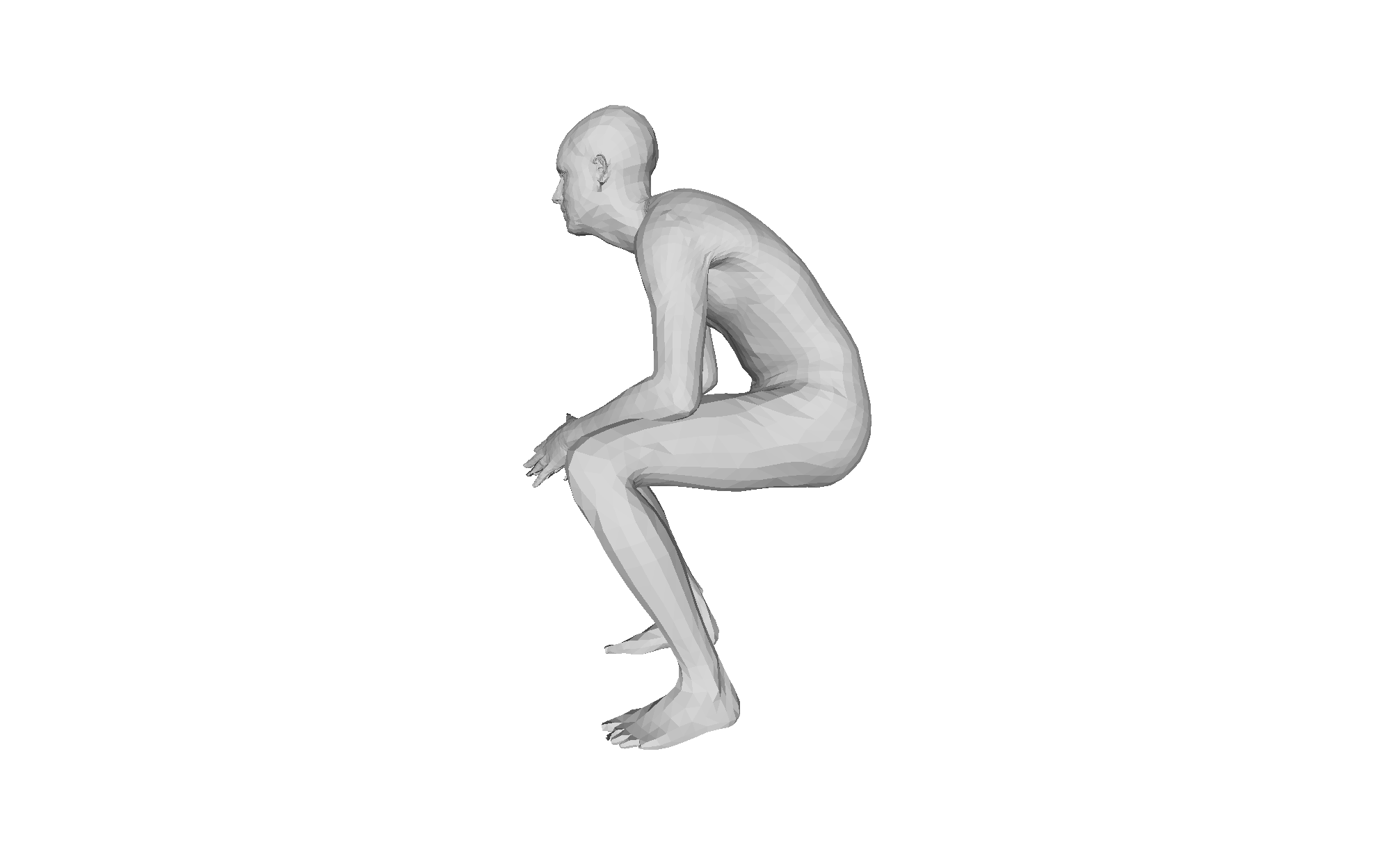}
            \includegraphics[width=0.19\textwidth,height=0.12\textwidth]{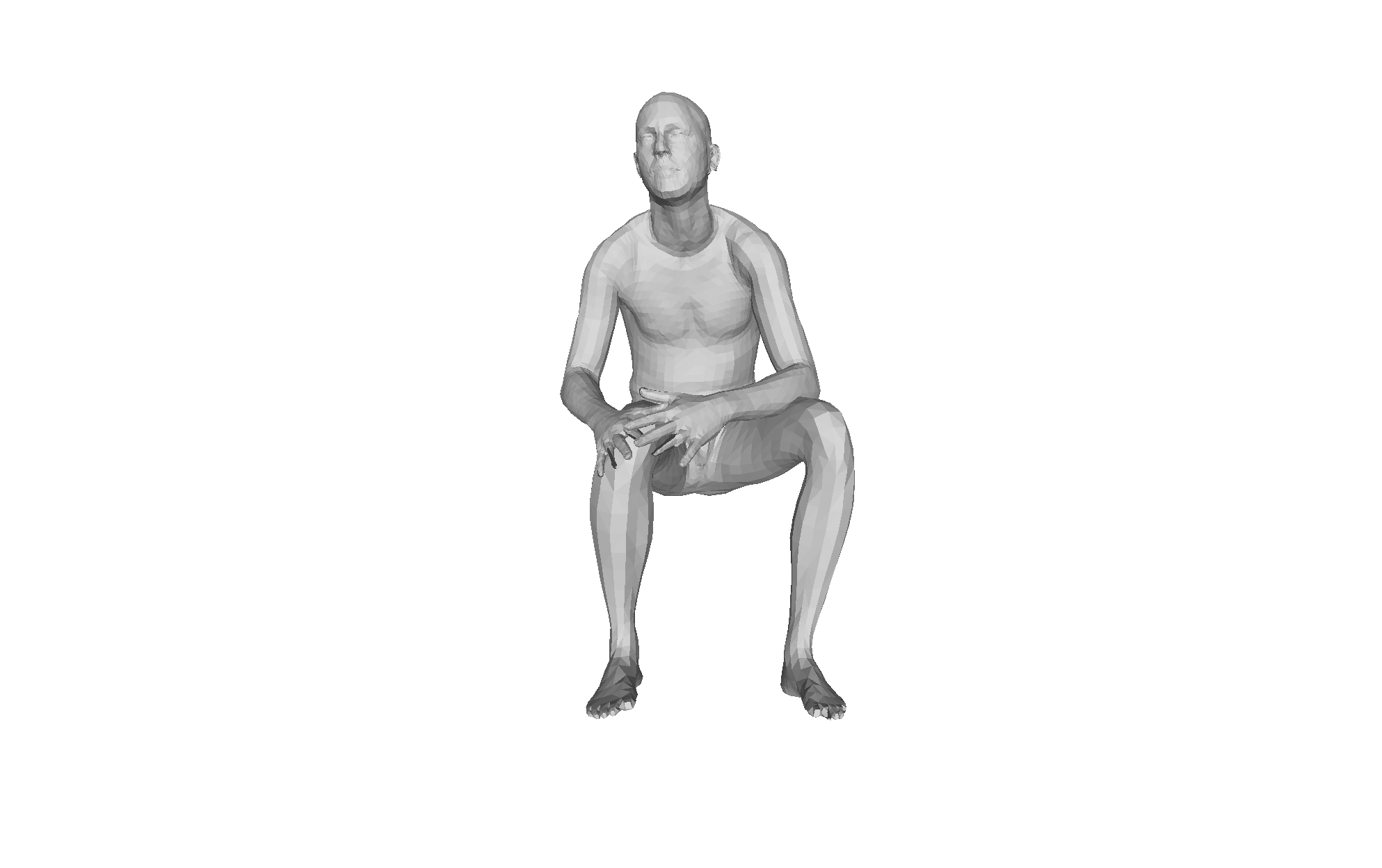}
            \\
            \includegraphics[width=0.19\textwidth,height=0.12\textwidth]{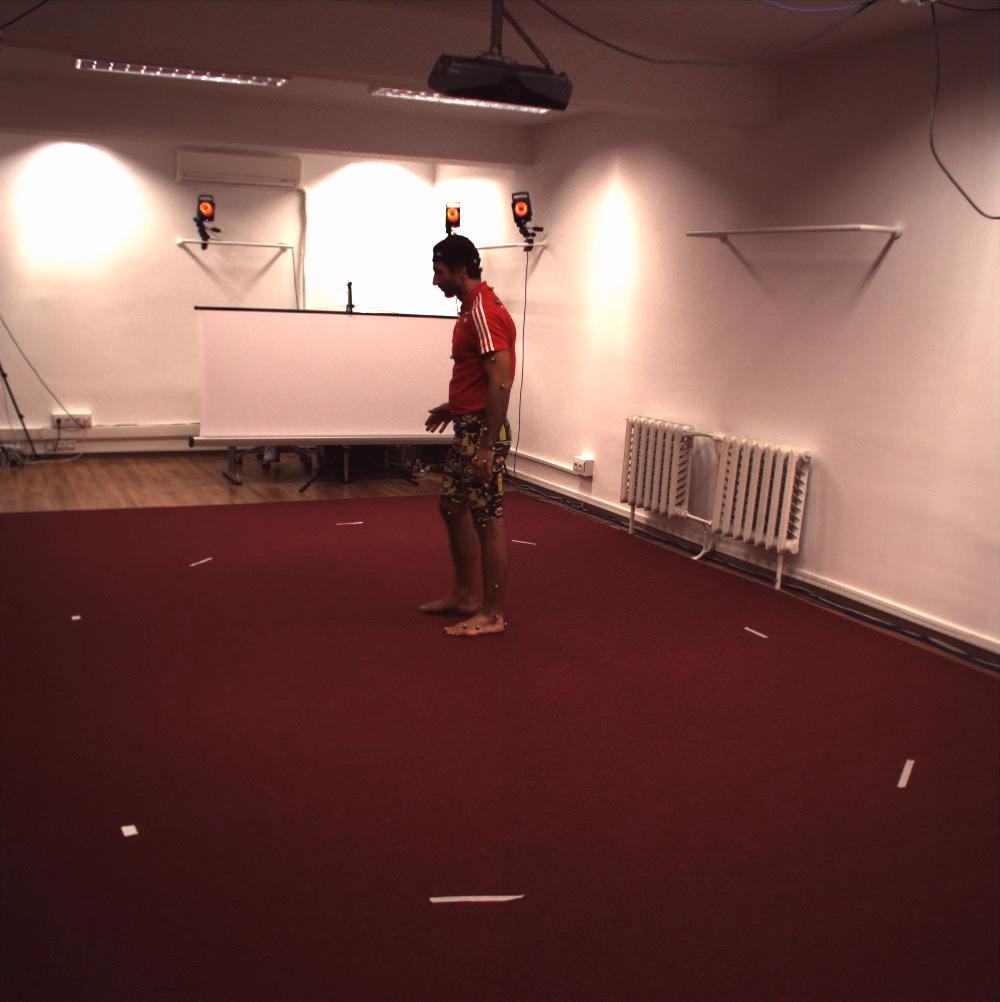}
            \includegraphics[width=0.19\textwidth,height=0.12\textwidth]{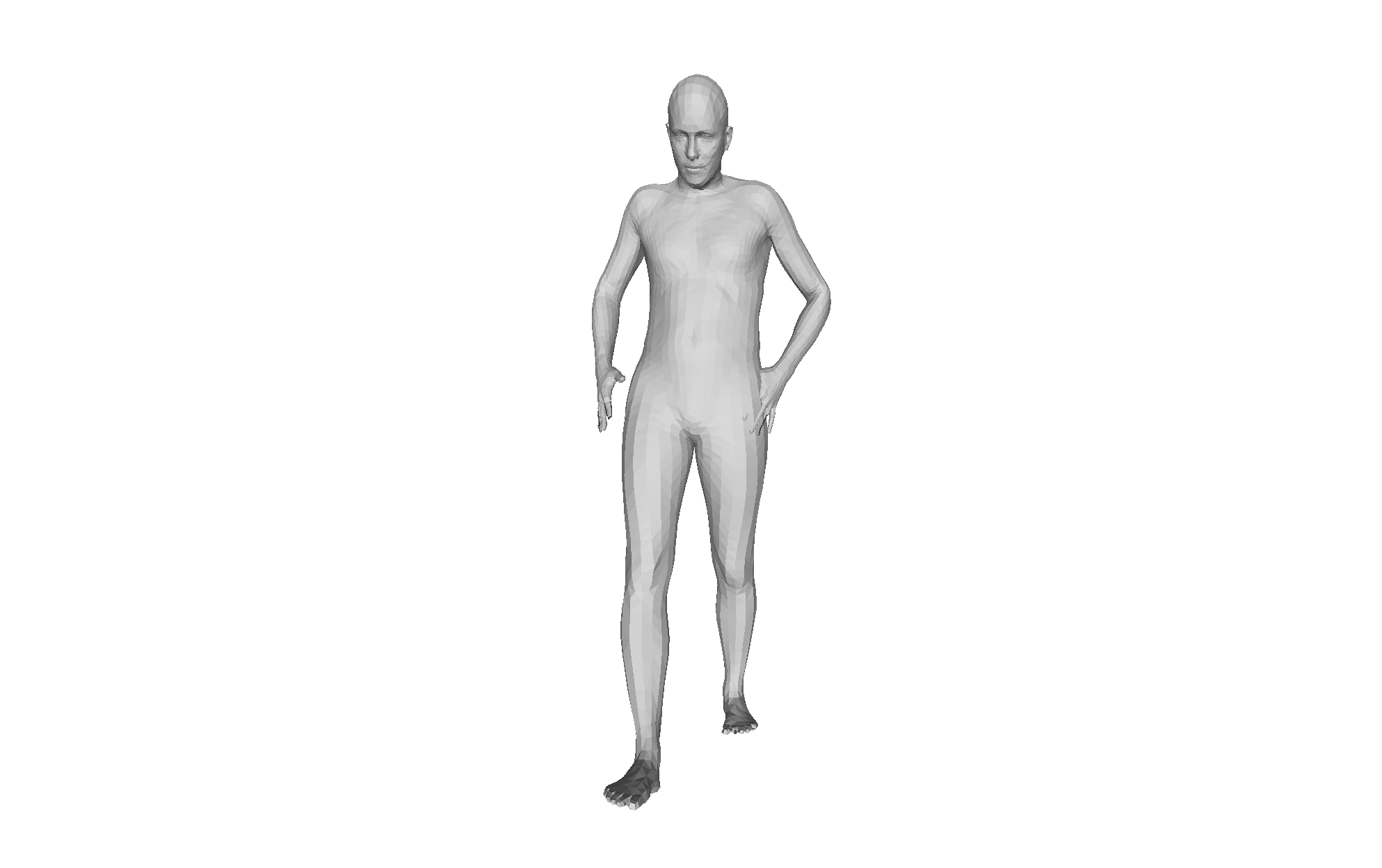}
            \includegraphics[width=0.19\textwidth,height=0.12\textwidth]{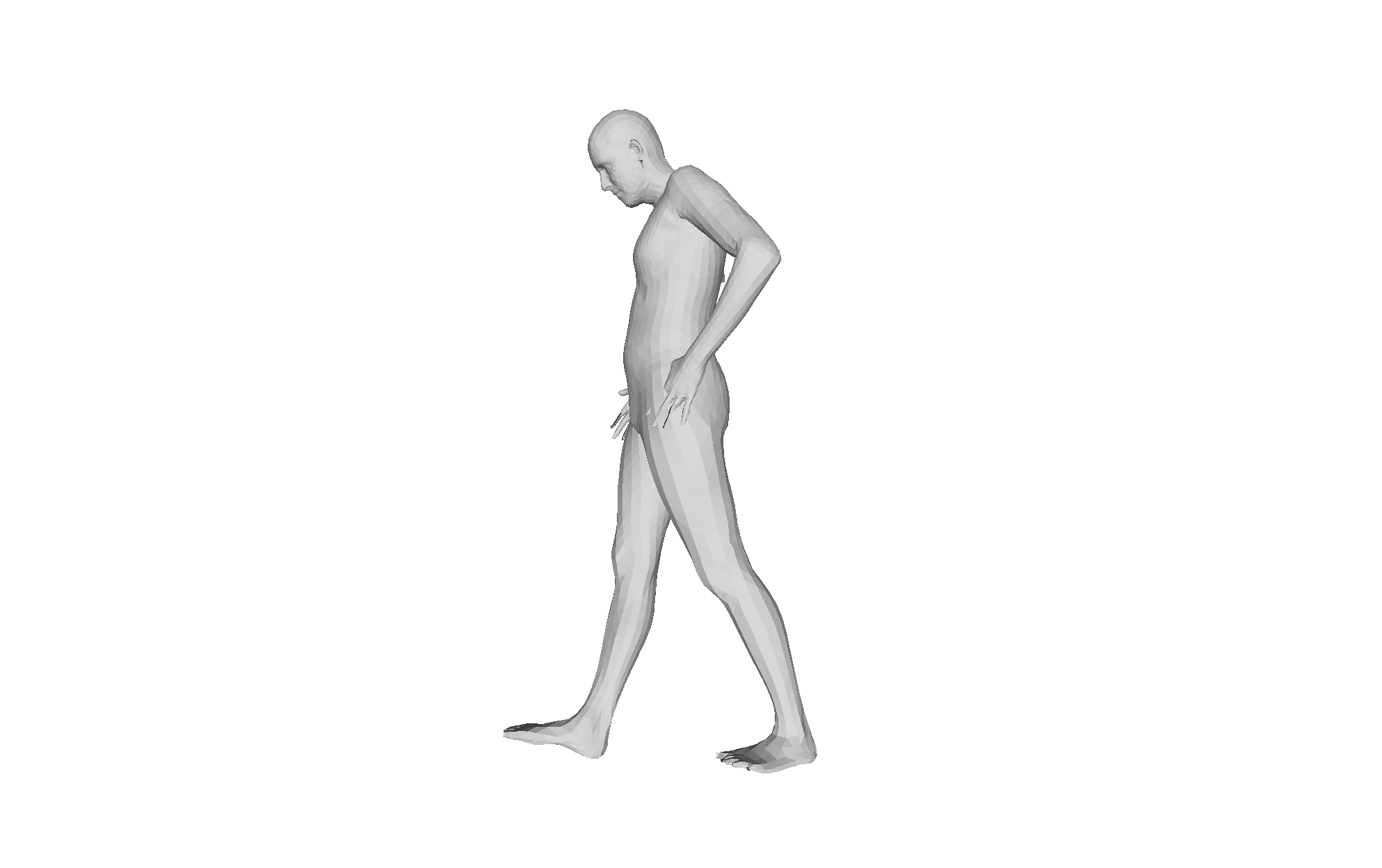}
            \includegraphics[width=0.19\textwidth,height=0.12\textwidth]{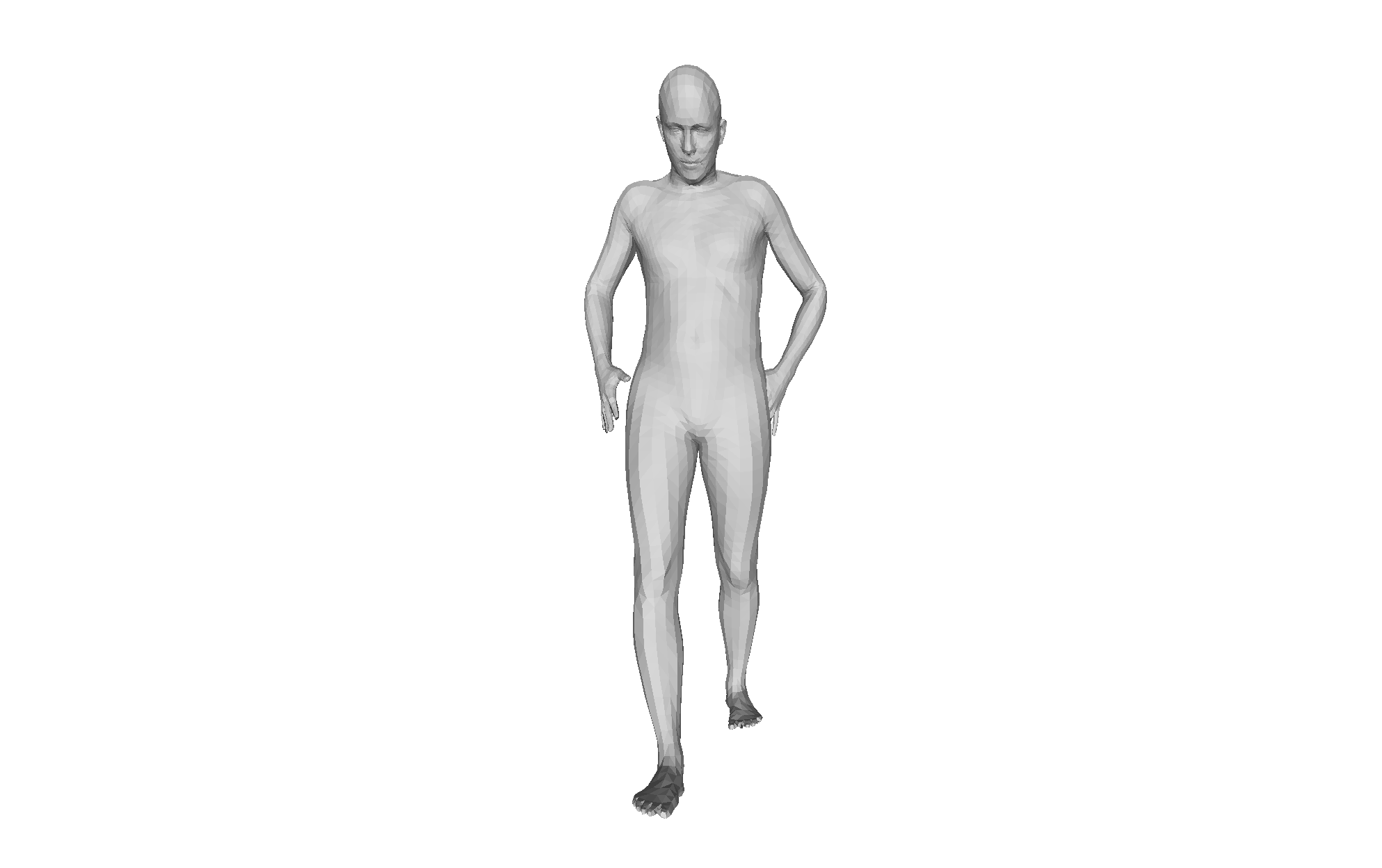}
            \includegraphics[width=0.19\textwidth,height=0.12\textwidth]{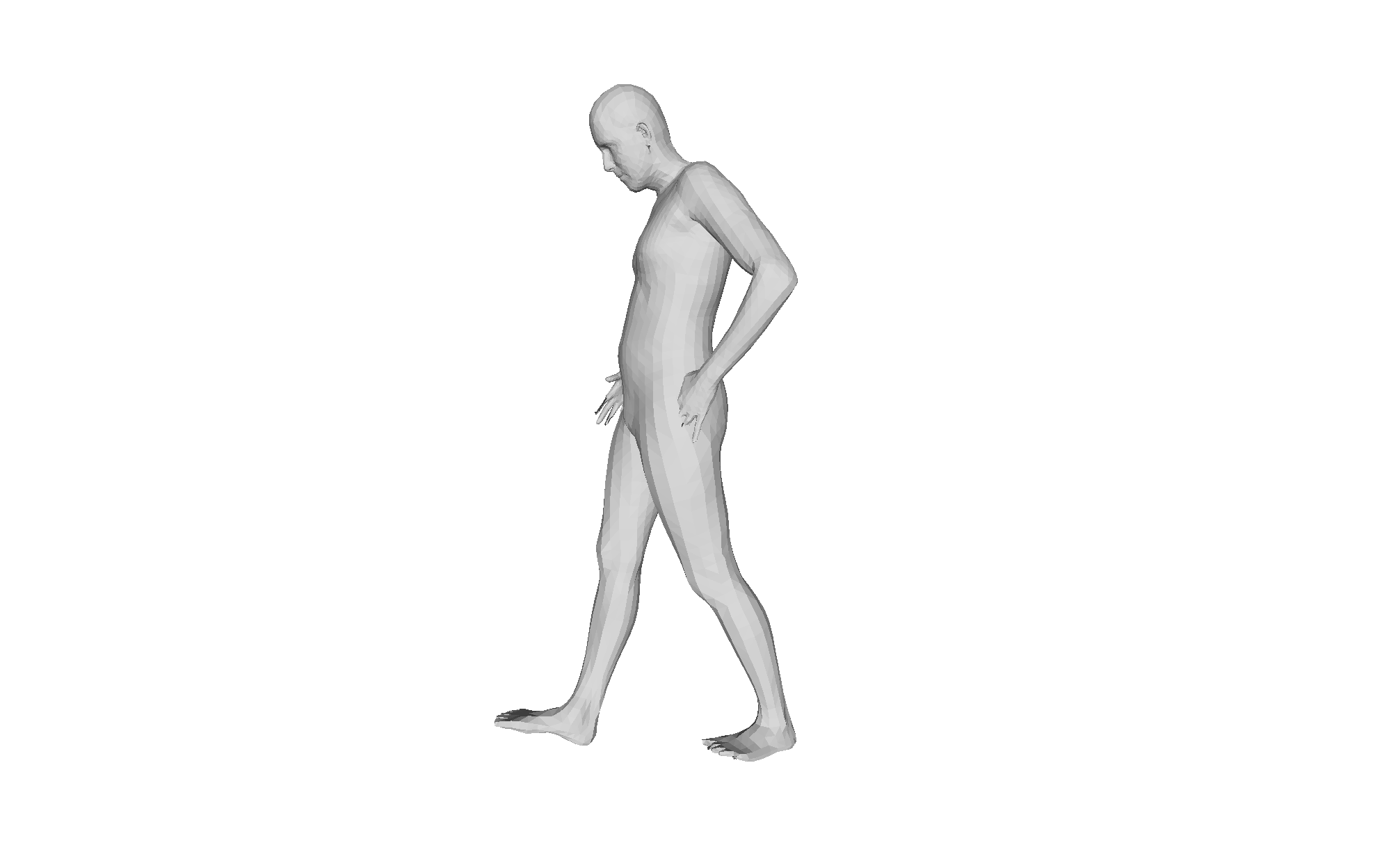}
            \\
            \includegraphics[width=0.19\textwidth,height=0.12\textwidth]{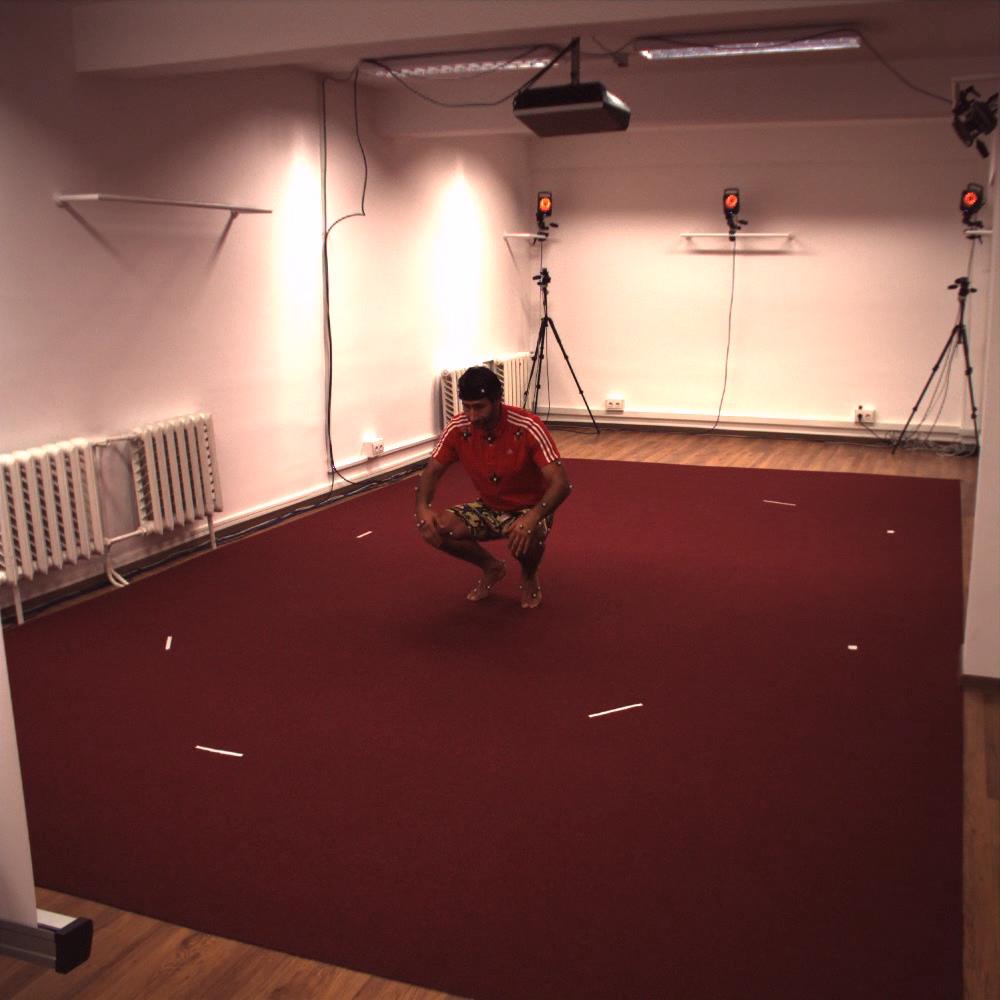}
            \includegraphics[width=0.19\textwidth,height=0.12\textwidth]{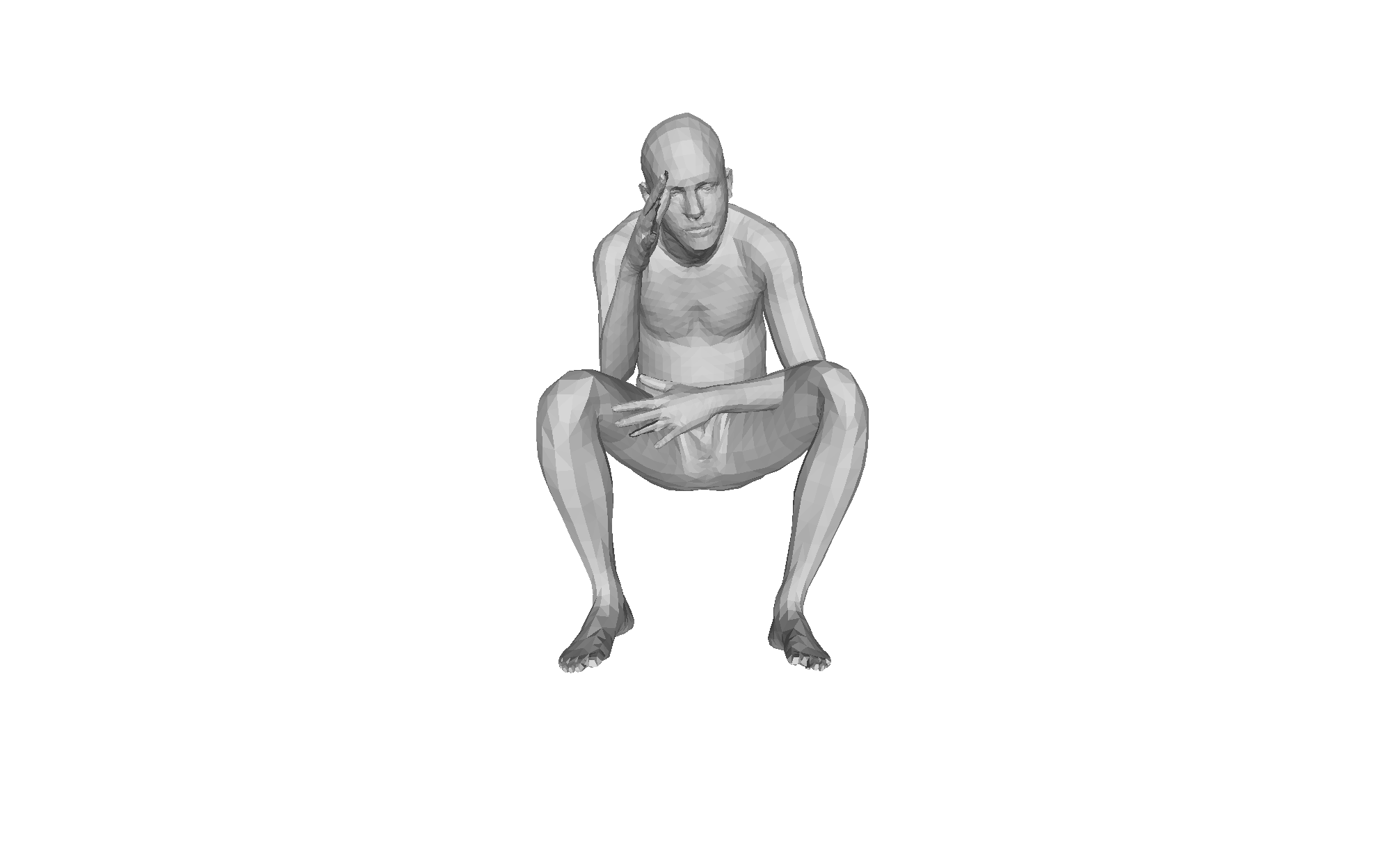}
            \includegraphics[width=0.19\textwidth,height=0.12\textwidth]{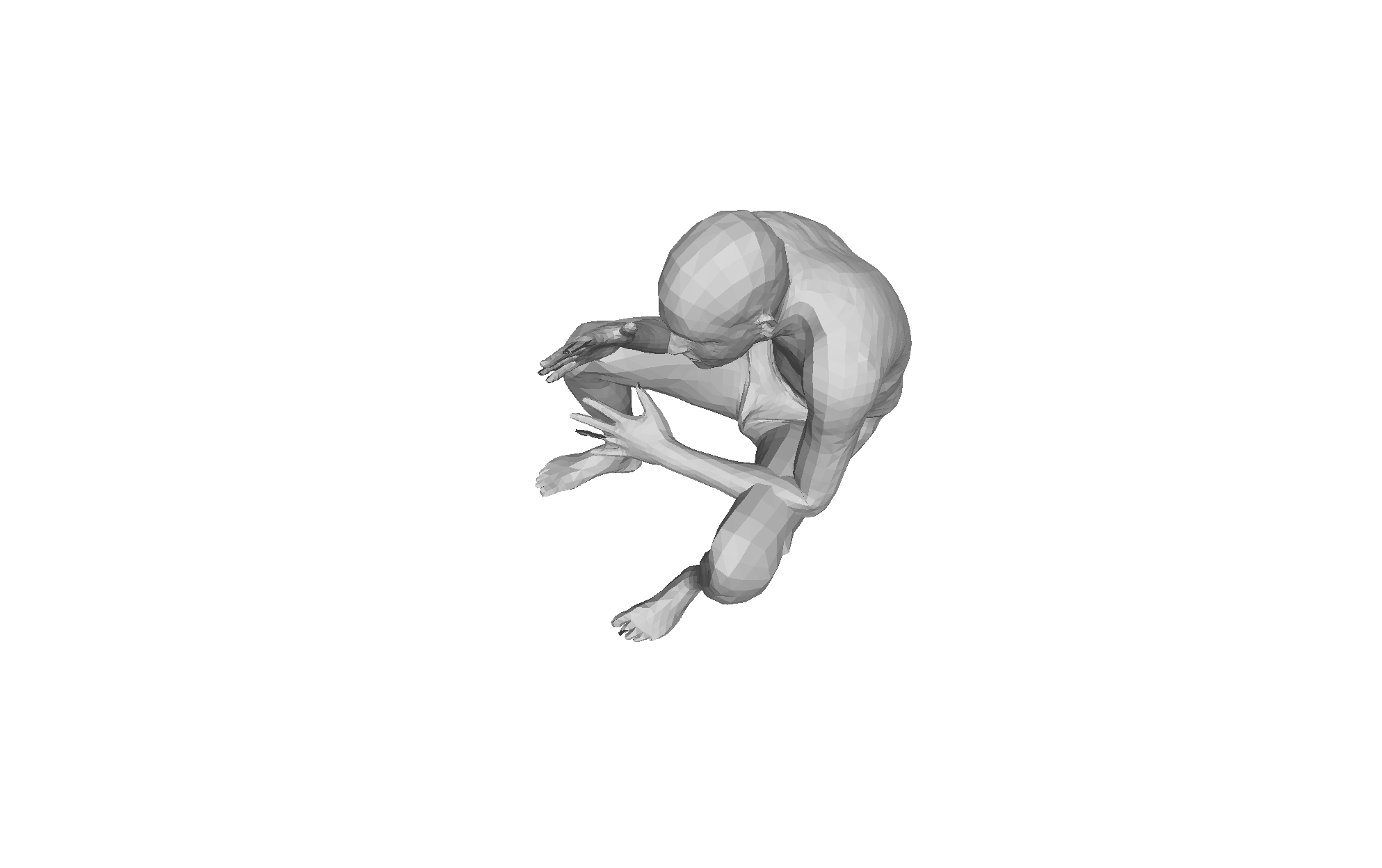}
            \includegraphics[width=0.19\textwidth,height=0.12\textwidth]{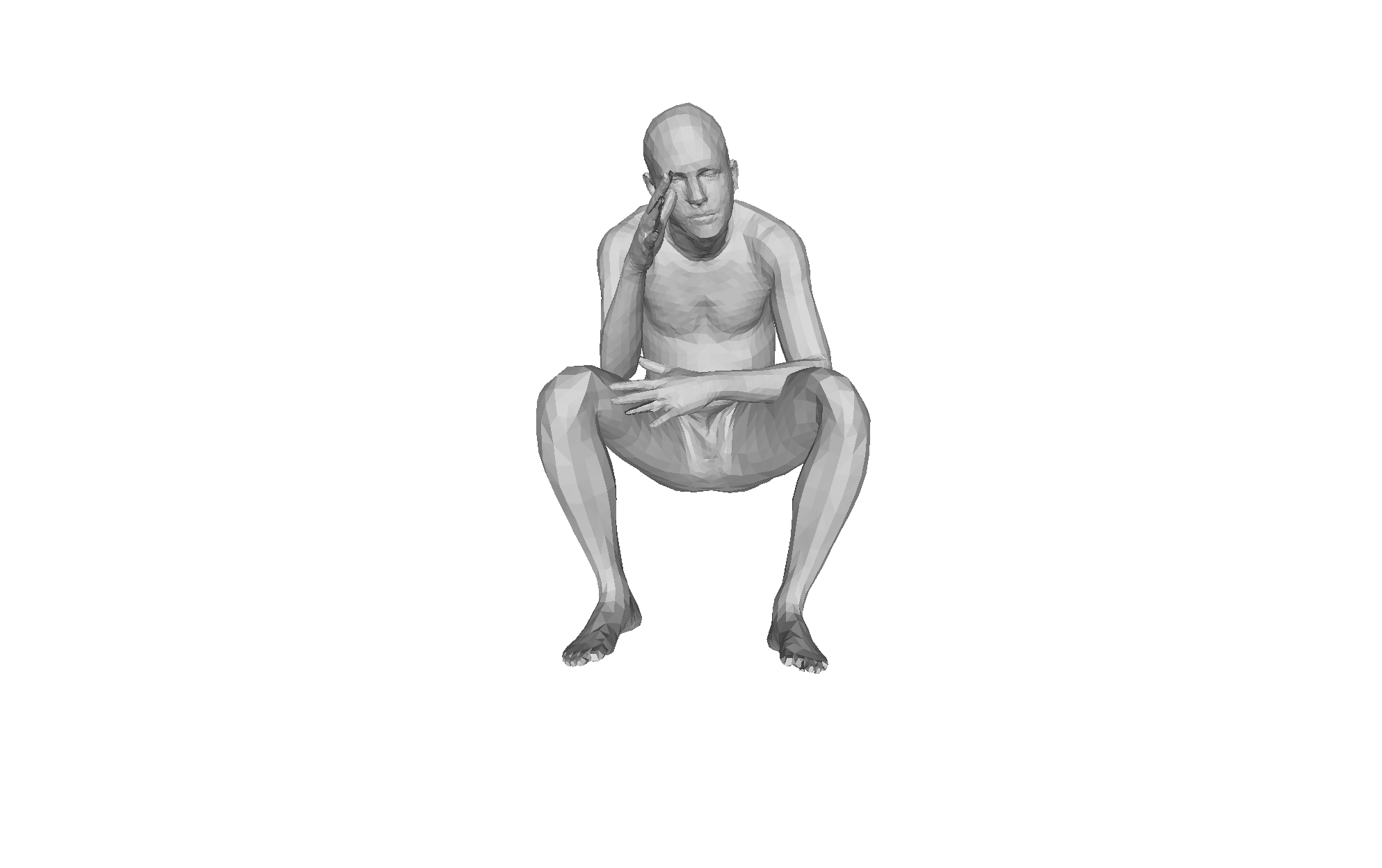}
            \includegraphics[width=0.19\textwidth,height=0.12\textwidth]{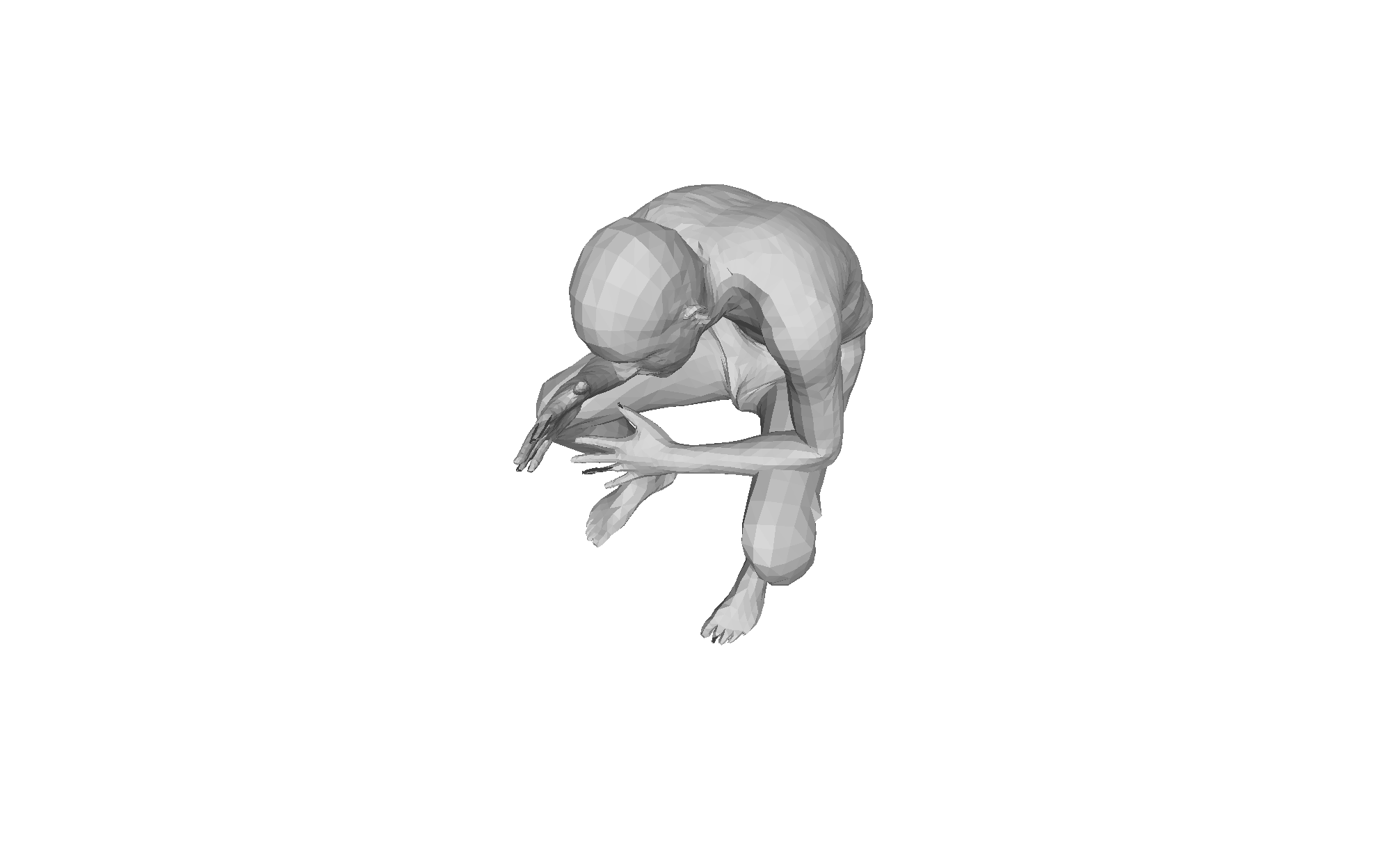}
            \\
            \includegraphics[width=0.19\textwidth,height=0.12\textwidth]{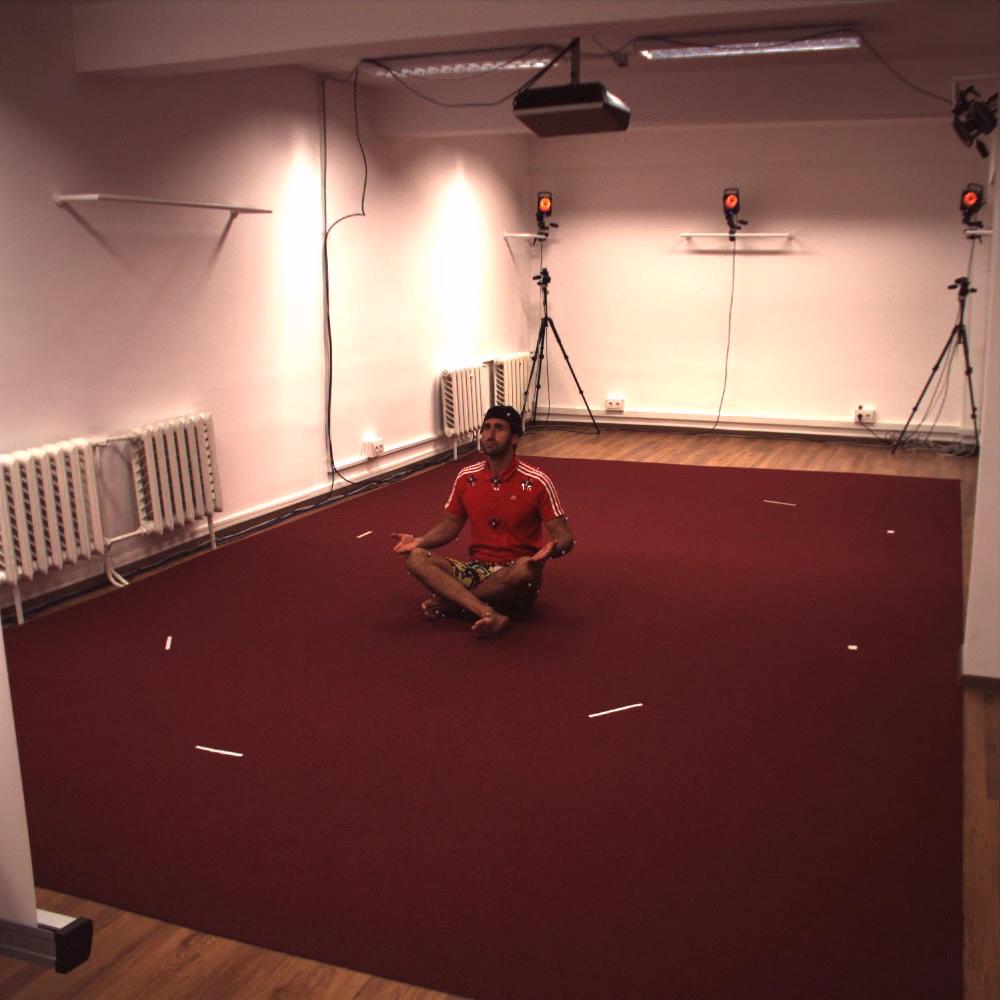}
            \includegraphics[width=0.19\textwidth,height=0.12\textwidth]{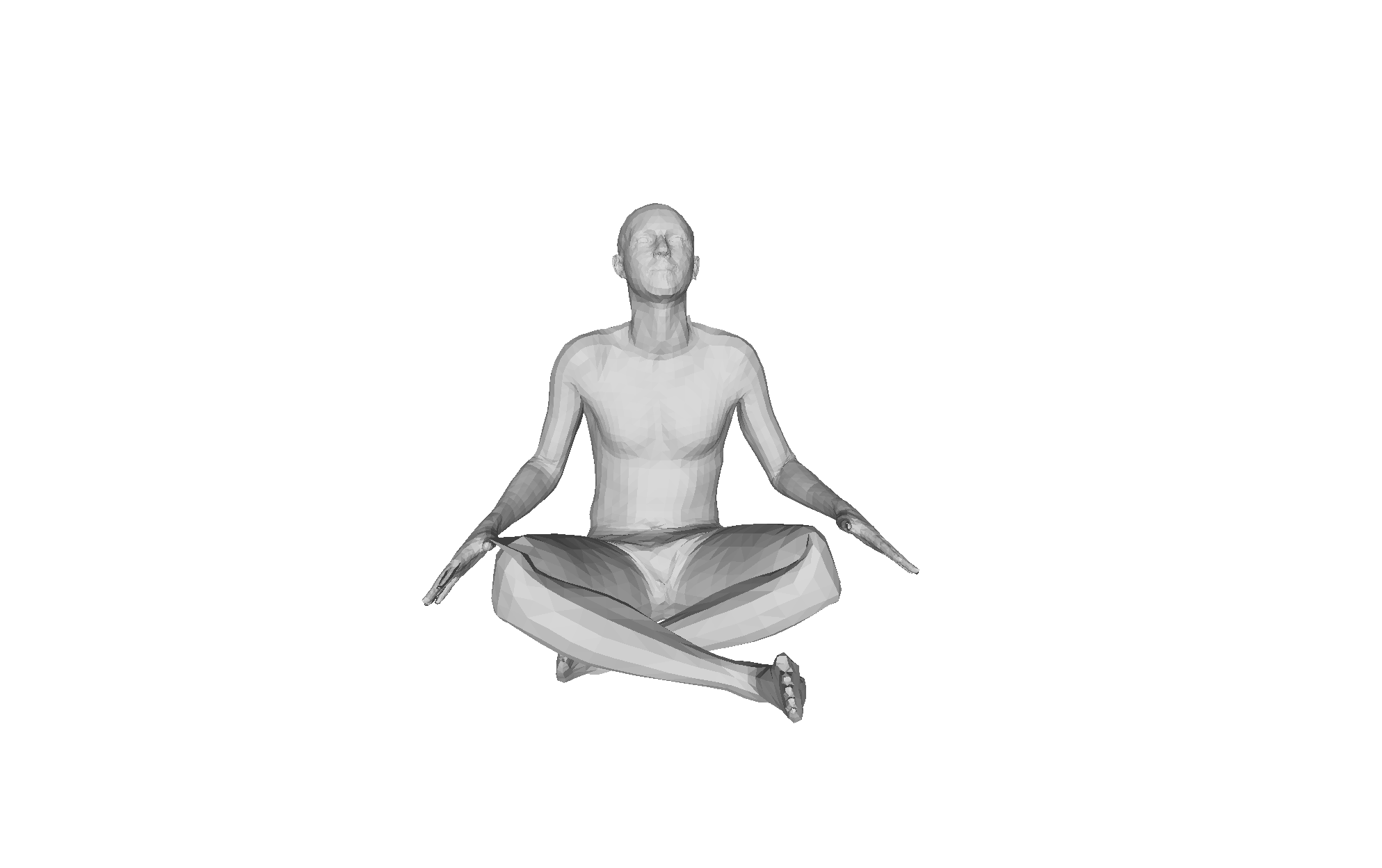}
            \includegraphics[width=0.19\textwidth,height=0.12\textwidth]{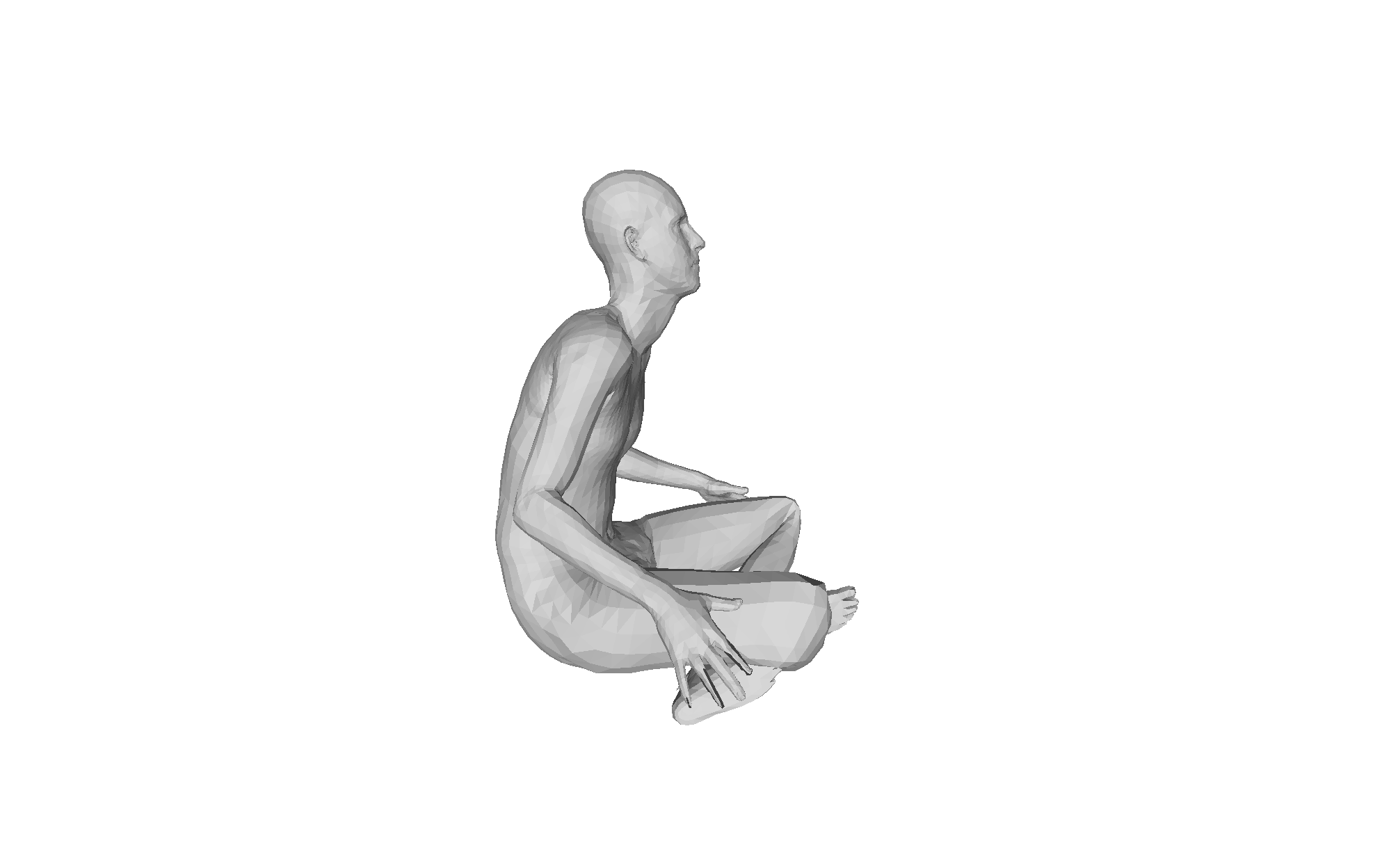}
            \includegraphics[width=0.19\textwidth,height=0.12\textwidth]{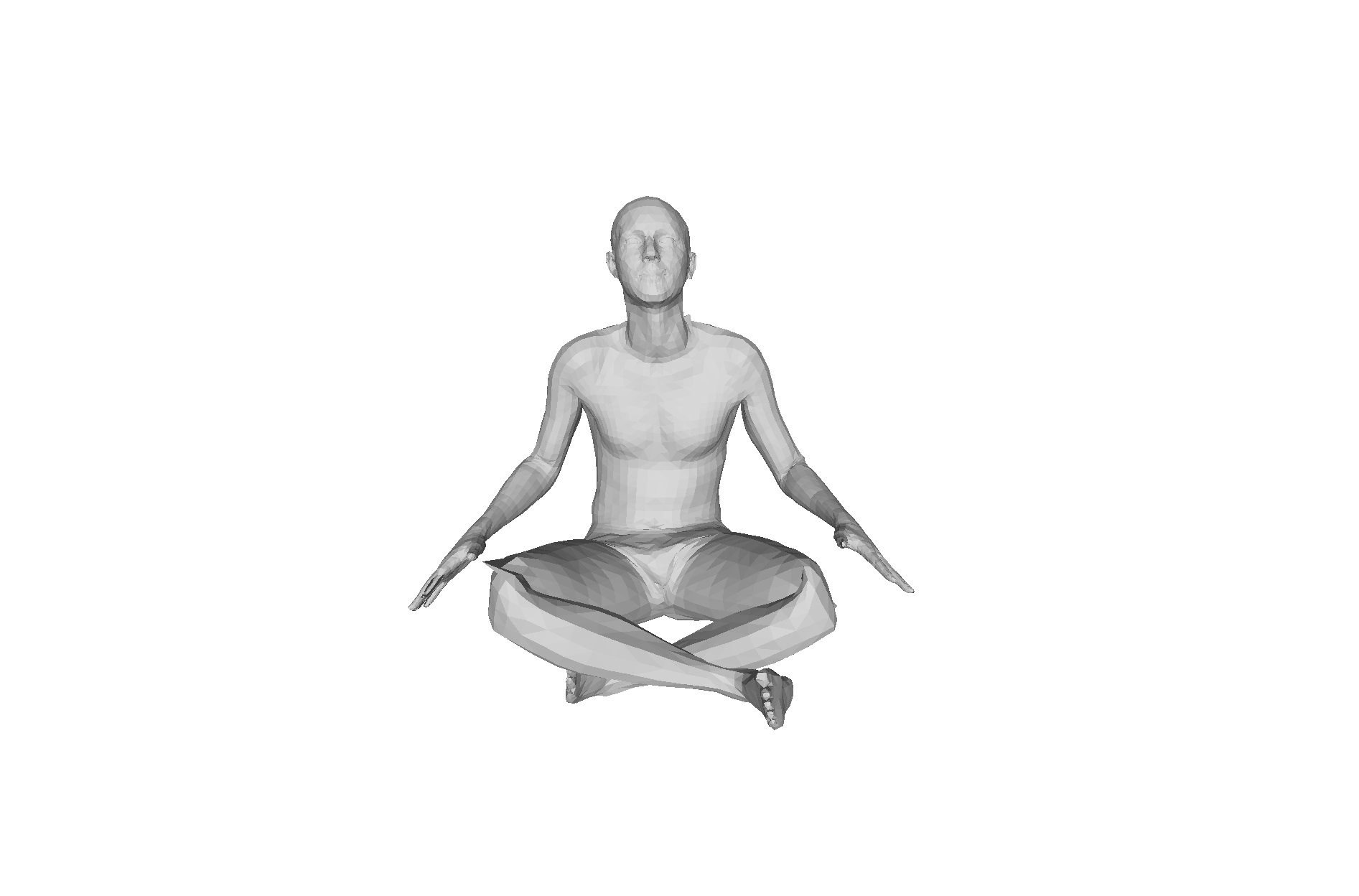}
            \includegraphics[width=0.19\textwidth,height=0.12\textwidth]{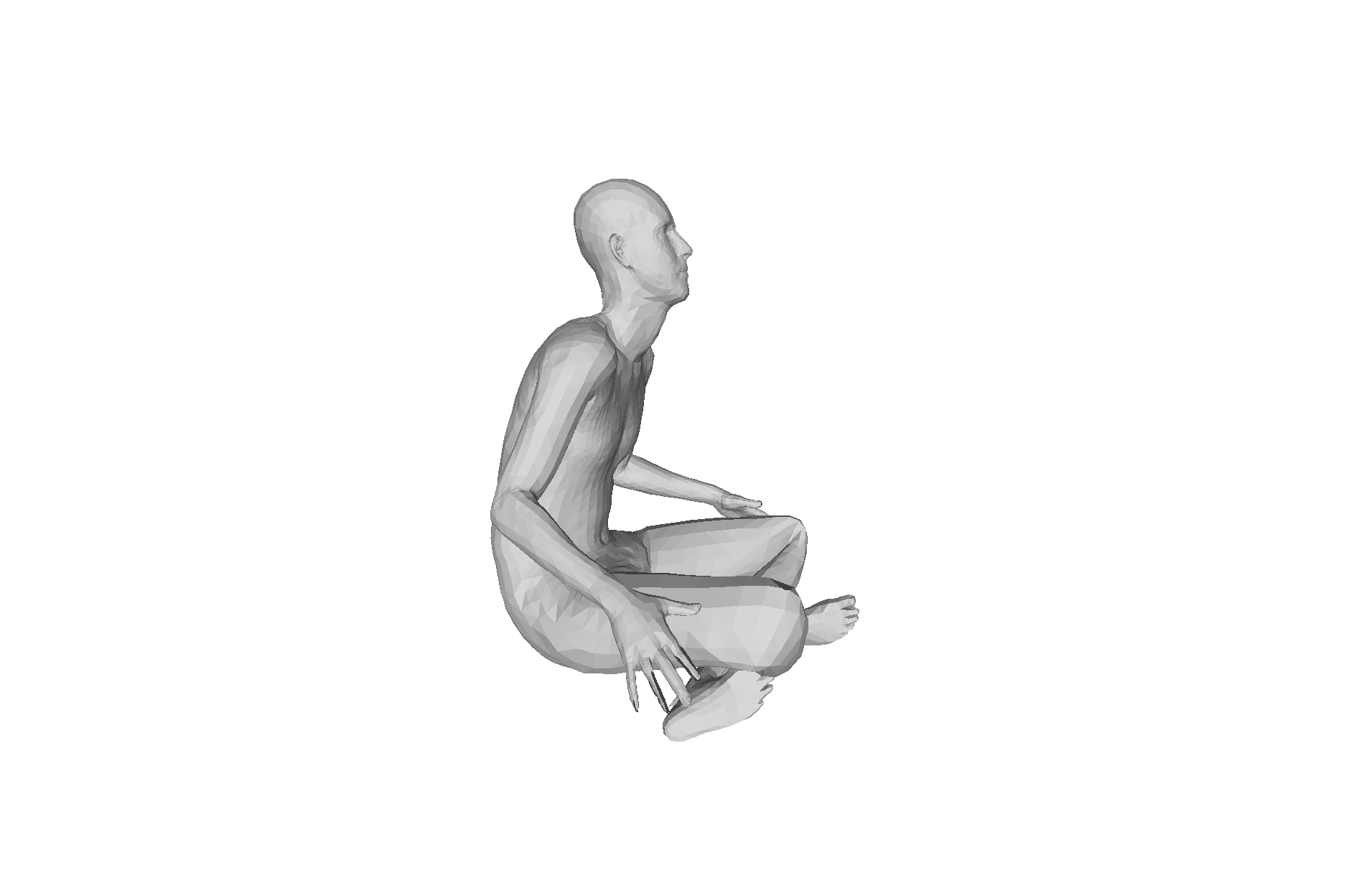}
            \\
            \parbox[c]{0.19\textwidth}{\centering sRGB}
            \parbox[c]{0.19\textwidth}{\centering ARTS-1}
            \parbox[c]{0.19\textwidth}{\centering ARTS-2}
            \parbox[c]{0.19\textwidth}{\centering HMRMamba-1}
            \parbox[c]{0.19\textwidth}{\centering HMRMamba-2}
        \end{minipage}
    }
	\caption{\textbf{Qualitative results on Human3.6m dataset.} HMR under occlusion from two viewpoints.}
	\label{fig:visual}
    \vspace{-2mm}
\end{figure*}

\noindent
In summary, quantitative results consistently demonstrate that HMRMamba excels across all benchmarks and the visualization is shown in~\cref{fig:visual}. Crucially, without introducing additional priors, it surpasses the previous SOTA ARTS, validating the efficacy of our proposed architecture.

\begin{table}
\centering
\caption{Ablation study for different 2D pose detectors on Human3.6M dataset.}
\vspace{-3mm}
\renewcommand\arraystretch{1.2}  
\small
\setlength{\tabcolsep}{0.8mm}{
\begin{tabular}{c|c|cccc}
\toprule
\multirow{2}{*}{Detector} & \multirow{2}{*}{Method}  & \multicolumn{4}{c}{Human3.6M}     \\  \cline{3-6} 
       & & MPJPE~$\downarrow$      & PA-MPJPE~$\downarrow$& MPVPE~$\downarrow$& Accel~$\downarrow$        \\ 
\hline \hline
\multirow{3}{*}{SH} & PMCE & 56.4 & 39.0 & 64.5 & \textbf{3.2} \\
                    & ARTS & 54.6 & 38.5 & 63.3 & 3.3 \\
                    & Ours & \textbf{52.9} & \textbf{37.3} & \textbf{61.3} & \textbf{3.2} \\
\hline \hline
\multirow{3}{*}{Detectron}  & PMCE   & 55.9 & 39.0 & 64.1 & \textbf{3.2} \\
                            & ARTS   & 53.7 & 38.1 & 62.8 & \textbf{3.2} \\
                            & Ours   & \textbf{52.4} & \textbf{37.9} & \textbf{61.8} & 3.3 \\
\hline \hline
\multirow{3}{*}{CPN}& PMCE  & 53.5 & 37.7& 61.3 & \textbf{3.1} \\
                    & ARTS & 51.6 & 36.6 & 60.2 & \textbf{3.1} \\
                    & Ours & \textbf{51.2} & \textbf{36.0} & \textbf{59.1} & \textbf{3.1} \\
\hline \hline
\multirow{3}{*}{GT} & PMCE & 36.3 & 26.8 & 46.2 & 2.2 \\
                    & ARTS & 33.7 & 24.8 & \textbf{45.7} & \textbf{2.0} \\
                    & Ours & \textbf{33.6} & \textbf{24.7} & 45.8 & 2.1 \\
\bottomrule
\end{tabular}
}
\label{table:detectors}
\vspace{-5mm}
\end{table}

\begin{table}
\centering
\caption{Ablation study for each component of HMRMamba on Human3.6M dataset.}
\vspace{-3mm}
\renewcommand\arraystretch{1.2}  
\small
\setlength{\tabcolsep}{0.7mm}{
\begin{tabular}{ccc|cccc}
\toprule
\quad GA\quad & \quad EM \quad & \quad IM \quad & MPJPE~$\downarrow$      & PA-MPJPE~$\downarrow$& MPVPE~$\downarrow$& Accel~$\downarrow$        \\ 
\hline \hline
\checkmark & & & 53.1 & 38.8 & 61.4 & 3.6 \\
\checkmark & \checkmark &  & 51.7 & 36.4 & 60.3 & 3.2\\
\checkmark &  & \checkmark & 52.3 & 36.9 & 61.9 & 3.1 \\
 & \checkmark & \checkmark & 52.4 & 37.1 & 61.6 & 3.3 \\
\checkmark & \checkmark & \checkmark & \textbf{51.2} & \textbf{36.0} & \textbf{59.1} & \textbf{3.1} \\
\bottomrule
\end{tabular}
}
\label{table:components}
\vspace{-5mm}
\end{table}

\begin{table}
\centering
\caption{Efficiency comparison of different models on Human3.6M dataset.}
\vspace{-3mm}
\renewcommand\arraystretch{1.2}  
\small
\setlength{\tabcolsep}{0.7mm}{
\begin{tabular}{c|c|c|cccc}
\toprule
Method & {\footnotesize \#P(M)} & {\footnotesize GFlops} & {\footnotesize MPJPE} & {\footnotesize PA-MPJPE} & {\footnotesize MPVPE} & {\footnotesize Accel} \\ 
\hline \hline
PMCE & 102.80 & 6.77 & 53.5 & 37.7& 61.3 & \textbf{3.1} \\
ARTS & 89.25  & 10.13 & 51.6 & 36.6 & 60.2 & \textbf{3.1} \\
Ours-S(vanilla) & 79.63  & 7.53 & 52.2 & 36.8 & 60.9 & 3.2 \\
Ours-S & 79.63  & 7.88 & \underline{51.2} & \underline{36.0} & \underline{60.2} & \underline{3.1} \\
Ours-L & 89.04  & 9.32 & \textbf{49.3} & \textbf{35.7} & \textbf{59.2} & \textbf{3.1} \\

\bottomrule
\end{tabular}
}
\label{table:efficiency}
\vspace{-1em}
\end{table}

\subsection{Ablation Study}
To validate our design, we conduct comprehensive ablation studies on Human3.6M, analyzing the impact of 2D pose detectors, the contribution of our core modules, and the model's computational efficiency.

\noindent \textbf{Impact of 2D Pose Detectors.}
We test HMRMamba's robustness using 2D poses from various detectors (SH, Detectron, CPN) and ground truth (GT). As shown in Table~\ref{table:detectors}, our method consistently outperforms PMCE and ARTS across all settings. With the CPN detector, HMRMamba achieves an MPJPE of \textbf{51.2~mm}, surpassing both competitors. This superiority indicates our Geometry-Aware Lifting Module is highly effective and resilient to detector noise. Its leading performance with GT poses (\textbf{33.6~mm} MPJPE) further confirms the fundamental strength of our architecture.

\noindent \textbf{Analysis of Model Components.}
We present an ablation analysis of HMRMamba’s components in Table~\ref{table:components}, focusing on the \textbf{Geometry-Alignment (GA)}, as well as the \textbf{Explicit (EM)} and \textbf{Implicit (IM)} motion representations. GA is the operation that leverages deformable attention to integrate spatial and temporal information from image features. As shown in the table, incorporating GA is critical for mesh reconstruction accuracy, resulting in an MPJPE improvement of approximately \textbf{1.2~mm}. Adding the motion modules (EM and IM) further enhances performance, particularly the motion coherence, with the full model achieving the best results: an MPJPE of \textbf{51.2~mm} and an Accel of \textbf{3.1~mm/s$^2$}. These results highlight the importance of combining a solid geometric foundation with comprehensive motion refinement for optimal mesh recovery.

\noindent \textbf{Efficiency Comparison.}
Table~\ref{table:efficiency} summarizes the efficiency of HMRMamba. Our model contains only \textbf{79.63M} parameters, making it 22.5\% smaller than PMCE and 10.8\% smaller than ARTS, which demonstrates the parameter efficiency of our Mamba-based architecture. In terms of computation, HMRMamba achieves \textbf{7.88 GFlops}, outperforming the Transformer-based ARTS and remaining close to PMCE, while providing a notable 4.3\% relative improvement in MPJPE. This reflects a favorable balance between accuracy and computational cost. Furthermore, compared to conventional bi-directional Mamba architectures, our STA-Mamba delivers significant accuracy improvements with only a marginal increase of 0.35 GFlops.


\vspace{-2mm}
\section{Conclusion}
\label{sec:conclusion}
Our code will be released soon.We introduced \textbf{HMRMamba}, a novel framework for video-based HMR that overcomes two critical limitations: the reliance on imperfect intermediate pose anchors and the inadequate modeling of spatiotemporal dynamics. Our approach tackles these flaws through two synergistic innovations. First, we establish a robust foundation with a \textbf{Geometry-Aware Lifting Module}, which leverages a unique dual-scan Mamba architecture to produce a reliable, geometrically-grounded 3D pose anchor sequence. Second, our \textbf{Motion-guided Reconstruction Network} injects crucial motion awareness by explicitly modeling the temporal evolution of the joint sequence, ensuring both physical plausibility and temporal coherence. Extensive evaluations demonstrate that HMRMamba establishes a new state-of-the-art with high efficiency.


{ \small
    \bibliographystyle{ieeenat_fullname}
    \bibliography{main}
}

\end{document}